\documentclass[sigconf]{acmart}

\copyrightyear{2026}
\acmYear{2026}
\setcopyright{cc}
\setcctype{by}
\acmConference[KDD '26]{Proceedings of the 32nd ACM SIGKDD Conference on Knowledge Discovery and Data Mining V.1}{August 09--13, 2026}{Jeju Island, Republic of Korea}
\acmBooktitle{Proceedings of the 32nd ACM SIGKDD Conference on Knowledge Discovery and Data Mining V.1 (KDD '26), August 09--13, 2026, Jeju Island, Republic of Korea}
\acmPrice{}
\acmDOI{10.1145/3770854.3780331}
\acmISBN{979-8-4007-2258-5/2026/08}

\usepackage{xcolor}

\usepackage{multirow}
\usepackage{enumitem}
\usepackage[table]{xcolor} % Required for coloring table cells
\usepackage[table]{xcolor} % Required for coloring table cells
\usepackage{graphicx}

\usepackage{amssymb}   % or amsmath/unicode-math

\usepackage{amsmath}
\usepackage{amssymb}
\usepackage{booktabs}
\usepackage{algorithm}
\usepackage{algorithmic}

\definecolor{lightblue}{RGB}{115, 246, 235} 
\definecolor{lightblue1}{RGB}{228, 255, 255} 
\usepackage{booktabs, makecell, xcolor, colortbl}

\definecolor{cvprblue}{rgb}{0.21,0.49,0.74}

\begin{document}

\title{Effective and Robust Multimodal Medical Image Analysis}

\author{Joy Dhar}
\orcid{0000-0001-9488-8298}
\affiliation{%
  \institution{Indian Institute of Technology Ropar}
  \city{Rupnagar}
  \state{Punjab}
  \country{India}
}\email{joy.22csz0003@iitrpr.ac.in}

\author{Nayyar Zaidi}
\affiliation{%
  \institution{Deakin University}
  \city{Melbourne}
  \state{Victoria}
  \country{Australia}}
\email{nayyar.zaidi@deakin.edu.au}

\author{Maryam Haghighat}
\affiliation{%
  \institution{Queensland University of Technology}
  \city{Brisbane}
  \state{QLD}
  \country{Australia}
}\email{maryam.haghighat@qut.edu.au}


%\renewcommand{\shortauthors}{Dhar et al.}


\begin{abstract}
\textbf{M}ultimodal \textbf{F}usion \textbf{L}earning (\texttt{MFL}), leveraging disparate data from various imaging modalities (e.g., \texttt{MRI}, \texttt{CT}, \texttt{SPECT}), has shown great potential for addressing medical problems such as skin cancer and brain tumor prediction. However, existing \texttt{MFL} methods face three key limitations: 
a) they often specialize in specific modalities, and overlook effective shared complementary information across diverse modalities, hence limiting their generalizability for multi-disease analysis;
b) they rely on computationally expensive models, restricting their applicability in resource-limited settings; and 
c) they lack robustness against adversarial attacks, compromising reliability in medical AI applications. 
To address these limitations, we propose a novel \textbf{M}ulti-\textbf{A}ttention \textbf{I}ntegration \textbf{L}earning (\texttt{MAIL}) network, incorporating two key components: a) an efficient
residual learning attention block for capturing refined modality-specific multi-scale patterns and b) an efficient multimodal cross-attention module for learning enriched complementary shared representations across diverse modalities. 
Furthermore, to ensure adversarial robustness, we extend \texttt{MAIL} network to design \texttt{Robust-MAIL} by incorporating random projection filters and modulated attention noise. 
Extensive evaluations on $20$ public datasets show that both \texttt{MAIL} and \texttt{Robust-MAIL} outperform existing methods, achieving performance gains of up to 9.34\% while reducing computational costs by up to 78.3\%. These results highlight the superiority of our approaches, ensuring more reliable predictions than top competitors.
\textcolor{blue}{Code:~\url{https://github.com/misti1203/MAIL-Robust-MAIL}.}
\end{abstract}

\begin{teaserfigure}
\centering
  \includegraphics[width=0.79\textwidth]{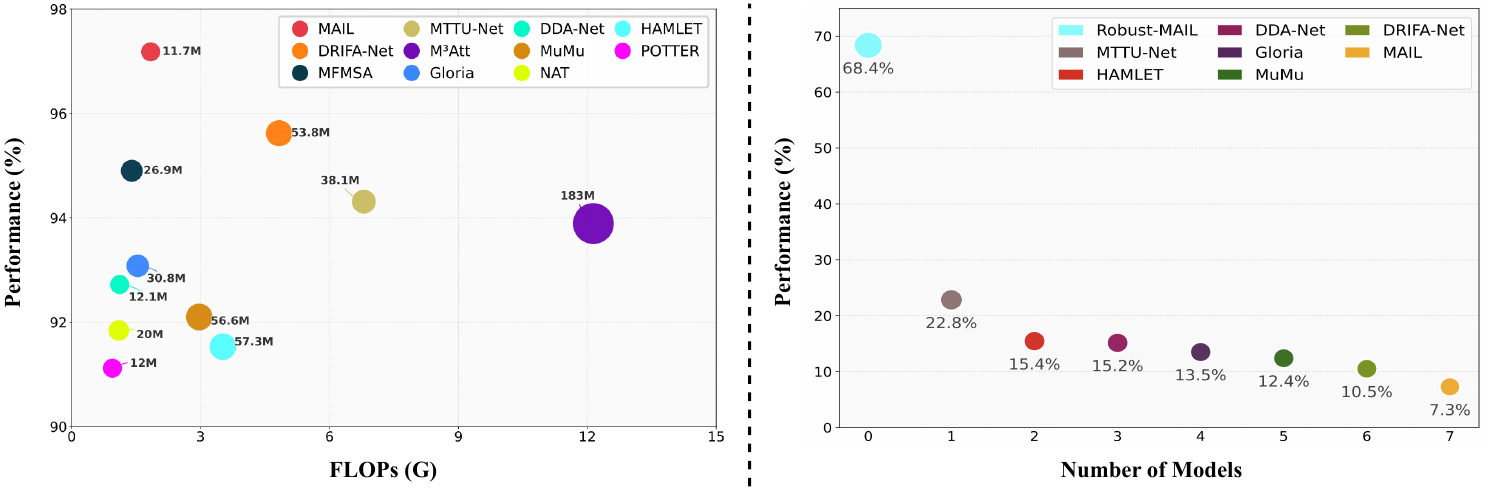}
      \vspace{-0.3cm}
  \caption{\small 
%\textbf{(A–B)} Attention‐alignment paradigms for multimodal fusion: cascaded attention (e.g., \texttt{DRIFA-Net}~\cite{dhar2024multimodal}, MuMu \cite{islam2022mumu}) vs. our parallel fusion attention (\texttt{EMCAM} in \texttt{MAIL}). The parallel design reduces information loss during shared‐representation learning unlike cascaded pipelines.  
\textbf{(Left)} \textbf{Cost–performance trade‐off:} comparison of our proposed \texttt{MAIL} framework vs.\ computationally-intensive fusion models (e.g., \texttt{DRIFA-Net} \cite{dhar2024multimodal}, \texttt{MuMu}~\cite{islam2022mumu}). 
\textbf{(Right)} \textbf{Adversarial robustness:}  comparison of proposed~\texttt{Robust-MAIL} vs.\ \texttt{SOTA} methods under a \texttt{PGD} attack.
}   
  \label{fig:fig1}
\end{teaserfigure}


\maketitle

%\vspace{-0.3cm}
%%%%%%%%%%%%%%%%%%%%%%%%%%%%%%%%%%%%%%%%%%%%%%%%%
\section{Introduction}
%%%%%%%%%%%%%%%%%%%%%%%%%%%%%%%%%%%%%%%%%%%%%%%%%

\begin{figure}[t]
    \centering
\includegraphics[width=0.45\textwidth]{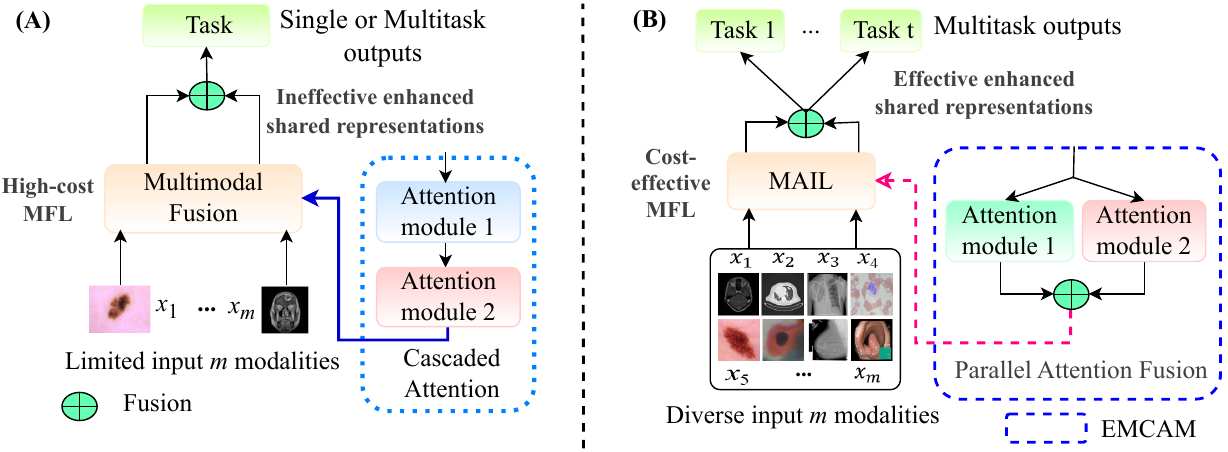}
        \vspace{-0.3cm}
    \caption{
\small \textbf{(A–B)} Attention‐alignment paradigms for \texttt{MFL}: cascaded attention (e.g., \texttt{DRIFA-Net}~\cite{dhar2024multimodal}, \texttt{MuMu} \cite{islam2022mumu}) vs. our parallel fusion attention (\texttt{EMCAM} in \texttt{MAIL}). The parallel design reduces information loss during shared‐representation learning unlike cascaded pipelines.}
    \label{fig:fig2}
    \vspace{-0.5cm}
\end{figure}

Many real-world applications rely fundamentally on multimodal data. A prime example is medical diagnostics where various modalities, such as magnetic resonance imaging (\texttt{MRI}), computed tomography scan (\texttt{CT-scan}), and chest x-ray (\texttt{CXR}), etc.\ play a crucial role in diagnosing critical health conditions, including brain tumors, various cancers, and other diseases. 
Recent advancements in deep learning for medical image analysis have revolutionized healthcare, offering rapid and cost-effective diagnostic solutions that improve decision-making for clinicians \cite{fitzgerald2022future}.
Existing methods \cite{dhar2024uncertainty,cui2023dual,  hassani2023neighborhood, zheng2023potter, rahman2024emcad} for handling multiple modalities, typically build models on a single modality and then apply strategies such as feature fusion and attention mechanisms to %\st{learn impactful features and} 
improve model performance. However, these approaches often result in suboptimal performance due to their lack of ability to effectively capture the most informative shared representations -- primarily because of noise present in medical data and issues with overfitting.

How to devise an effective model that is capable of leveraging various modalities – also known as Multimodal Fusion Learning (\texttt{MFL}), remains an open challenge in the machine learning community. \texttt{MFL} learns information from multiple modalities to capture enhanced shared representations, thereby improving predictive performance \cite{dhar2024multimodal, islam2020hamlet, islam2022mumu}. These methods address key limitations faced by the above-mentioned single-modal learning models by capturing shared complementary representations from diverse modalities. However, these methods still face performance limitations due to their restricted ability to capture enriched representations, mainly because they do not incorporate attention mechanisms. 
In recent years, attention-based models have gained popularity for their ability to automatically learn the importance of individual tokens (i.e., elements of interest) \cite{vaswani2017attention}. Their application to \texttt{MFL} has shown promise, with several notable attention-based \texttt{MFL} approaches ~\cite{islam2020hamlet, huang2021gloria, cheng2022fully} being developed to enrich representations and improve performance across medical imaging modalities.

% -----------------------------------------------
%\vspace{-0.3cm}
%\begin{figure}[t]
 %   \centering
%\includegraphics[width=0.37\textwidth]{fig_2.pdf}
 %   \vspace{-0.4cm}
  %  \caption{
%\small \textbf{(A)} Adversarial-attack pipeline revealing model vulnerabilities. \textbf{(B)} Robust-MAIL workflow that mitigates these vulnerabilities and achieve adversarial robustness.}
 %   \label{fig:fig2}
  %  \vspace{-0.6cm}
%\end{figure}

% -----------------------------------------------
Despite the success of \texttt{MFL} methods, existing approaches usually face four fundamental challenges. First, compared to single-modal learning networks, existing \texttt{MFL} incurs \textbf{significantly higher computational costs} (\textit{\textcolor{blue}{challenge 1}}) due to computationally intensive convolutions or attention modules, posing substantial efficiency barriers that must be overcome for practical applications.
Second, existing \texttt{MFL}s often process attention modules in a cascaded manner (e.g., sequentially stacking modules), which potentially \textbf{risks progressive information loss during transitions between modules} (\textit{\textcolor{blue}{challenge 2}}). This design choice can significantly degrade the retention of important representations.
Third, current \texttt{MFL} approaches have \textbf{limited capacity to effectively learn shared complementary representations} (\textit{\textcolor{blue}{challenge 3}}) across diverse medical imaging modalities, regardless of whether they employ single attention mechanisms or cascaded multi-attention modules. 
This is because, their reliance on disease-specific modalities -- such as \texttt{MRI}, \texttt{PET}, \texttt{CT}, \texttt{SPECT}, dermoscopy, or Pap smears (targeting brain disorders, skin/lung/cervical cancers) -- during representation learning process, potentially \textbf{limits their generalizability} for multi-disease classification.
Fourth, although \texttt{MFL} methods achieve notable performance improvements, they remain \textbf{vulnerable to adversarial attacks} (\textit{\textcolor{blue}{challenge 4}}), potentially limiting their reliability in AI-driven healthcare applications. Even {minor adversarial perturbations} can result in {incorrect predictions} directly effecting patient outcomes. Without robust adversarial defenses, such systems may generate unreliable diagnoses, posing serious risks to patient safety.

In this paper, we focus on \emph{addressing cost-effectiveness, information loss in cascaded modules, limited generalization due to restricted modalities, and adversarial vulnerability} -- four fundamental challenges in state-of-the-art (\texttt{SOTA}) \texttt{MFL} methods. To address these concerns, we propose two groundbreaking frameworks -- \textbf{M}ulti-\textbf{A}ttention \textbf{I}ntegration \textbf{L}earning (\textbf{MAIL}) and \texttt{Robust-MAIL}-jointly redefining AI-driven medical imaging applications. These frameworks achieve \emph{high performance under computational constraints} while ensuring \emph{reliability against adversarial attacks}, as shown in Figs.~\ref{fig:fig1}-\ref{fig:fig2}. 
MAIL tackles the \textit{\textcolor{blue}{first three challenges}} through two core modules: \textbf{E}fficient \textbf{R}esidual \textbf{L}earning \textbf{A}ttention (\texttt{ERLA}) and \textbf{E}fficient \textbf{M}ultimodal \textbf{C}ross-\textbf{A}ttention \textbf{M}odule (\texttt{EMCAM}). Specifically, \texttt{ERLA} \textit{\textcolor{blue}{addresses challenge 1}} by refining multi-scale channel dependencies through low-cost convolutions with channel attention, enhancing representational diversity while preserving computational efficiency (ref. Tables \ref{tab:tab1}-\ref{tab:tab2}, \ref{tab:tab4}-\ref{tab:tab5}). 
\texttt{EMCAM} \textit{\textcolor{blue}{solves challenges 1-2}} by pioneering parallelized fusion of multimodal information across frequency and spatial domains, optimizing performance-efficiency trade-offs while minimizing information loss (ref. Tables \ref{tab:tab1}-\ref{tab:tab2}, \ref{tab:tab6}). 

\begin{algorithm}[H]
\renewcommand{\thealgorithm}{1}
\caption{\texttt{MAIL} Network}
\label{al:al1}
\begin{algorithmic}[1]
\setlength{\itemsep}{1pt}
\setlength{\parskip}{1pt}
\setlength{\baselineskip}{14pt}
\setlength{\jot}{1pt}
\setlength{\abovedisplayskip}{1pt}
\setlength{\belowdisplayskip}{1pt}
\STATE \textbf{Input:} $m$-th multimodal input $x_i \in X = [x_1, x_2, \ldots, x_m]$, where $x_{[1:m]} \in \mathbb{R}^{H \times W \times C}$
\STATE \textbf{Output:} Multi-disease tasks (e.g., classification) via two salient phases: \texttt{MSTL} and \texttt{TMTL}
\STATE \textbf{Procedure:}
\STATE \textit{/* \texttt{MAIL} network with the proposed \texttt{ERLA} and \texttt{EMCAM} blocks. Base network: \texttt{ResNet18} */}
\IF{phase == \texttt{MSTL} (Modality-Specific Task Learning)}
    \FOR{each modality $i \in [1:m]$}
        \STATE Apply a $7 \times 7$ convolutional layer as the input layer: \\
        \hspace{2em}$x_i \gets \text{Conv}(x_i)$
        \FOR{each filter size in \{64, 128, 256, 512\}}
            \STATE Refine multi-scale diverse patterns using \texttt{ERLA} block (Eqs. 1--4): \\
            \hspace{2em}$x^{\prime}_{i} \gets \texttt{ERLA}(x_i)$
            \STATE Learn enhanced complementary shared representations using \texttt{EMCAM} block (Eqs. 5--13): \\
            \hspace{2em}$X^S \gets \texttt{EMCAM}(x^{\prime}_{i})$
        \ENDFOR
    \ENDFOR
\ENDIF
\IF{phase == \texttt{TMTL} (Target-Specific Multitask Learning)}
    \STATE Perform multi-task learning:
    \STATE \small{$\mathcal{L}_{\texttt{TMTL}} = \sum_{t=1}^{T} \sum_{m=1}^{M} \lambda_t^M \cdot \mathcal{L}_t^M \left( \mathcal{F}(X^S; \beta), Y \right), \quad$}
    \STATE $\beta^* = \arg \min_\beta \mathcal{L}_{\texttt{TMTL}}$
\ENDIF

\end{algorithmic}
\end{algorithm}

\begin{figure*}[ht!]
    \centering
    \includegraphics[width=\textwidth]{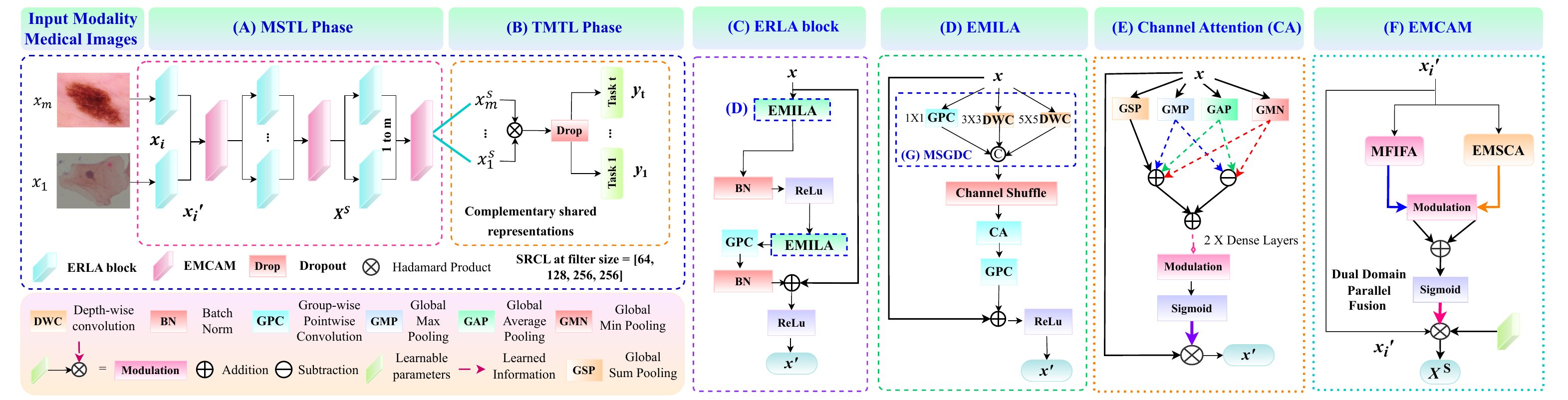}
        \vspace{-0.4cm}
    %\caption{\small Detailed architecture of \texttt{MAIL} network. Key components include: (A) the \texttt{MSTL} phase captures enhanced complementary shared representations, while (B) \texttt{TMTL} phase facilitates multi-disease classification. \texttt{MSTL} integrates (C) Efficient Residual Learning Attention (\texttt{ERLA}) block to learn refined multi-scaled patterns and (F) Efficient Multimodal Cross Attention Module (\texttt{EMCAM}) comprises: Multimodal Frequency-domain Information Fusion Attention (\texttt{MFIFA}) and the Efficient Multimodal Spatial-domain Cross Attention (\texttt{EMSCA}) for joint optimization of global frequency contexts and spatial dependencies. \texttt{ERLA} incorporates (D) the \texttt{EMILA} module, comprising (D) Multi-Scale Group with Depth-wise Convolutions (\texttt{MSGDC}), channel shuffle, and (E) Channel Attention (\texttt{CA}), to refine modality-specific multi-scaled representations.}
    \caption{\small Architecture of \texttt{MAIL} network (comprising of (A) \texttt{MSTL} and (B) \texttt{TMTL} phases). Key components with-in~\texttt{MSTL} phase are (C) \texttt{ERLA} and (F) \texttt{EMCAM} blocks.
    \texttt{ERLA} is based on (D) \texttt{EMILA} module, comprising of (G) \texttt{MSGDC} block inspired from \texttt{EMCAD's} \texttt{MSDC} ~\cite{rahman2024emcad} and (E) Channel Attention (\texttt{CA}). \texttt{EMCAM} details are in Figure~\ref{fig:fig4}.
    %\texttt{MFIFA} and \texttt{EMSCA} for joint optimization of global frequency contexts and spatial dependencies in parallel. , to refine multi-scaled representations.
    }
    \label{fig:fig3}
    \vspace{-0.2cm}
\end{figure*}

To \textit{\textcolor{blue}{deal with challenge 3}}, \texttt{MAIL} learns from and adapts to diverse medical imaging modalities, enhancing its generalizability for multi-disease classification (see Tables \ref{tab:tab1}-\ref{tab:tab2}). To \textit{\textcolor{blue}{mitigate challenge 4}}, we extend \texttt{MAIL} to \texttt{Robust-MAIL}, ensuring reliable predictions under adversarial attacks. We first explore the random projection filter \cite{dong2023adversarial} and observe its effectiveness in enhancing \texttt{MFL} robustness (ref. Tables \ref{tab:tab3}-\ref{tab:tab5}). Motivated by this, we redesign \texttt{MAIL} into \texttt{Robust-MAIL} by integrating \textbf{R}andom \textbf{P}rojection with \textbf{A}ttention \textbf{N}oise (\texttt{RPAN}), which combines random projection filtering with modulated attention noise. \texttt{RPAN} \textit{injects stochasticity} into \texttt{ERLA} and \texttt{EMCAM}, disrupting adversarial perturbations while preserving crucial information, transforming \texttt{MAIL} into a robust framework.  
%We highlight the main contributions of our work as follows: 
Our main contributions can be summarized as follows:
%\vspace{-0.5cm}
\begin{itemize}[leftmargin=*]
\item[$\bullet$] We introduce the \texttt{MAIL} network, designed to enhance \texttt{MFL}'s effectiveness by jointly optimizing frequency and spatial domain information through parallel fusion. This enables effective multimodal information fusion achieving optimal performance while minimizing computational cost. 
%\item[$\bullet$] We design the \texttt{ERLA} block for refined modality-specific multi-scale representation learning via incorporating the Efficient Multi-Scale Information Learning Attention (\texttt{EMILA}) module. 
%\item[$\bullet$] We design \texttt{EMCAM} block for enhanced complementary shared representation learning. \texttt{EMCAM} comprises: (a) Multimodal Frequency-Domain Information Fusion Attention (\texttt{MFIFA}) module for capturing multi-frequency multimodal global contexts and (b) Efficient Multimodal Spatial-Domain Cross Attention (\texttt{EMSCA}) module for refining multimodal spatial details. These modules fuse frequency and spatial domain information in parallel, capturing global-spatial multimodal contexts across diverse modalities. 
\item[$\bullet$] We introduce \texttt{Robust-MAIL}, an extension of \texttt{MAIL} designed for adversarial robustness. It integrates random projection filters with modulated attention noise mechanisms to ensure reliable predictions against adversarial attacks.
\item[$\bullet$] We conduct extensive evaluations against state-of-the-art methods, demonstrating significant improvements across 20 diverse medical imaging datasets. %\textcolor{red}{A summary of these results is given in Figure~\ref{fig:fig2}.}
\end{itemize}

%\vspace{-0.3cm}
% 

%%%%%%%%%%%%%%%%%%%%%%%%%%%%%%%%%%%%%%%%%%%%%%%%%%%%%%%%%%%%%%%
%\vspace{-0.5cm}
\section{Related Study} \label{sec:rel}
% %%%%%%%%%%%%%%%%%%%%%%%%%%%%%%%%%%%%%%%%%%%%%%%%%%%%%%%%%%%%%%%
Previous works in \texttt{MFL} have been widely explored in natural vision tasks \cite{peng2022balanced, wang2020learning, joze2020mmtm, ma2021smil}, with limited adaptation to medical imaging \cite{cheng2022fully}. For example, \texttt{MTTU-Net} \cite{cheng2022fully} combined \texttt{CNN}s and transformers for glioma segmentation and IDH genotyping but lacked explicit attention mechanisms to learn nuanced multimodal representations.
Recent attention-based architectures, including \texttt{DDA-Net} \cite{cui2023dual}, \texttt{MADGNet} \cite{nam2024modality}, \texttt{NAT} \cite{hassani2023neighborhood}, \texttt{POTTER} \cite{zheng2023potter}, and \texttt{EMCAD} \cite{rahman2024emcad}, have demonstrated significant advancements in representation learning for both natural and medical imaging tasks.
The integration of attention mechanisms into \texttt{MFL} frameworks has played a pivotal role in further enhancing their performance \cite{islam2020hamlet, islam2022mumu, song2024low, huang2021gloria, sun2022event, zhou2023cacfnet, He2023, liu2023multi}.
For instance, \texttt{GLORIA} \cite{huang2021gloria} leveraged global-local attention to align radiology reports with images for label-efficient learning. \texttt{HAMLET} \cite{islam2020hamlet} and \texttt{MuMu} \cite{islam2022mumu} employed multi-head and lightweight self-attention, respectively, to fuse multimodal sensor data for human activity recognition. 
In contrast, \texttt{M$^3$Att} \cite{liu2023multi} used mutual attention to integrate features from dual modalities, enhancing segmentation performance. Similarly, \texttt{CAF} \cite{He2023} and \texttt{DRIFA-Net} \cite{dhar2024multimodal} adopted cross-modal and cascaded dual attention for skin cancer diagnosis and multi-disease classification, highlighting the versatility of attention mechanisms.
However, as described earlier, current \texttt{MFL} methods often suffer from computational inefficiency due to computationally intensive convolutions or attentions, potentially risk progressive information loss from cascaded attention transitions, and limited adaptability to learn shared complementary representations effectively. 
These studies do not utilize the benefits of lightweight multi-scale depth-wise convolutions, parallelized attention fusion to mitigate information loss, and adaptability to diverse medical imaging modalities for enhanced generalizability in multi-disease classification. 
Therefore, in this work, we propose \texttt{ERLA} and \texttt{EMCAM}, integrated within our \texttt{MAIL} network, to enhance model performance while maintaining computational feasibility -- by integrating lightweight multi-scale depth-wise convolutions and parallelized attention mechanism.
\texttt{MFL} networks must ensure reliable predictions against adversarial examples, necessitating the development of robust~\texttt{MFL} architectures. While several defenses -- \texttt{PNI} \cite{he2019parametric}, \texttt{DBN} \cite{han2021advancing}, \texttt{Learn2Perturb} \cite{jeddi2020learn2perturb}, \texttt{RP} filter \cite{dong2023adversarial}, \texttt{CAP} \cite{xiang2023toward}, and \texttt{CTRW} \cite{ma2024adversarial} -- exist, they primarily focus on adversarial defense in single-modal networks and are often designed for natural images. These approaches overlook the challenges posed by medical imaging, where significant differences in data size, feature characteristics, and task patterns complicate the direct application of natural vision defenses to medical imaging applications \cite{Dong2023AdversarialAttackDefense}. To bridge this gap, we extend \texttt{MAIL} with a random projection filter and modulated attention noise, forming \texttt{Robust-MAIL} to %enhance adversarial robustness and 
ensure reliable medical AI applications.

\begin{figure*}[ht!]
    \centering
    \includegraphics[width=\textwidth]{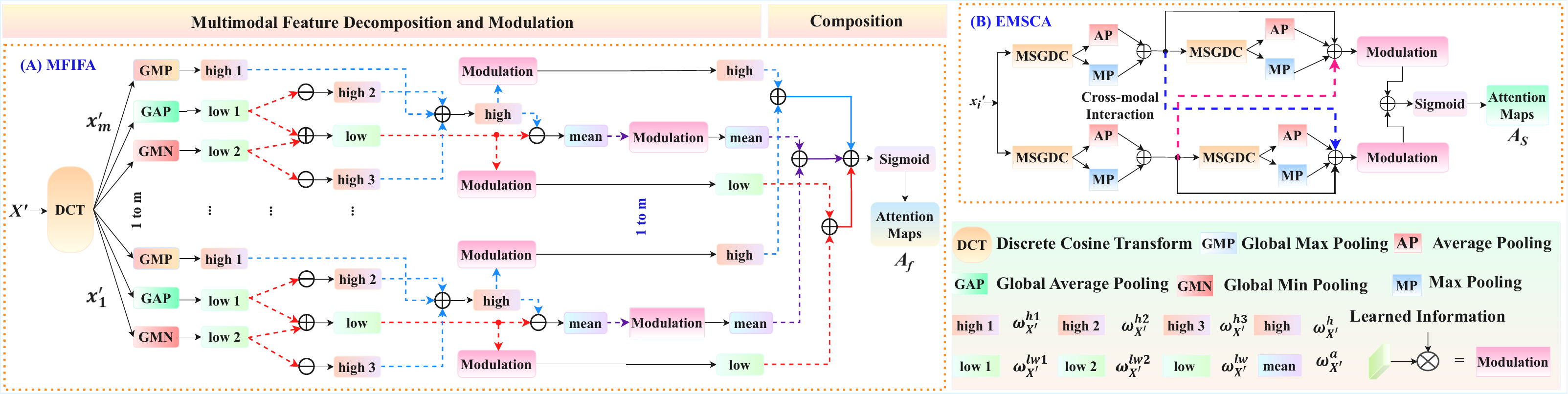}
        \vspace{-0.4cm}
    \caption{
    \small Components of \texttt{EMCAM}: (A) \texttt{MFIFA} module captures multi-frequency multimodal global contexts, while (B) \texttt{EMSCA} module refines multimodal spatial representations. 
    }
    \label{fig:fig4}
    \vspace{-0.3cm}
\end{figure*}

%\vspace{-0.6cm}
%%%%%%%%%%%%%%%%%%%%%%%%%%%%%%%%%%%%%%%%%%%%%%%%%%%%%%%%%%%%%%%
\section{Proposed Method}
%%%%%%%%%%%%%%%%%%%%%%%%%%%%%%%%%%%%%%%%%%%%%%%%%%%%%%%%%%%%%%%
%\noindent\textbf{Problem Formulation: } In this section, let us discuss our proposed effective multimodal fusion learning network: \texttt{MAIL} for multimodal image analysis. Given an input \( X = [x_1, \dots, x_m] \) from \( m \) diverse modalities and corresponding labels \( Y = [y_1, \dots, y_t] \) for \( t \) disease classification tasks, where \( y_t \in \{0, 1, \dots, n\} \) represents binary or multi-class labels, \texttt{MAIL}, denoted as \( \mathcal{F}(\cdot) \), aims to learn enhanced complementary shared representations \( X^S = [x_1^s, \dots, x_m^s] \) while mapping \( \mathcal{F}(X) \to Y \), aiming to balance optimal performance with minimal computational cost.
\noindent\textbf{Problem Formulation: } In this section, let us discuss our proposed  multimodal fusion learning network: \texttt{MAIL}, for multimodal image analysis. Given an input \( X = [x_1, \dots, x_m] \) from \( m \) diverse modalities and labels \( Y = [y_1, \dots, y_t] \) for \( t \) disease classification tasks, \texttt{MAIL}, denoted as \( \mathcal{F}(\cdot) \), aims to learn enhanced complementary shared representations \( X^S = [x_1^s, \dots, x_m^s] \) while optimizing \( \mathcal{F}(X) \to Y \) for improved performance with minimal computational cost.

\noindent\textbf{Method Overview: }
We provide a holistic overview of our proposed \texttt{MAIL} network, as shown in Fig.~\ref{fig:fig3} (A, B). \emph{Detailed steps of \texttt{MAIL} are provided in \textcolor{blue}{Algorithm~\ref{al:al1}}}. We will show how \texttt{MAIL} utilizes information fusion learning with efficient attention strategies, ensuring adaptability across diverse neural architectures while maintaining optimal performance with minimal computational overhead. It comprises of two salient phases: \textit{Modality-Specific Task Learning (\texttt{MSTL}) and Target-Specific Multitask Learning (\texttt{TMTL}). \texttt{MSTL} captures enhanced complementary shared representations, while \texttt{TMTL} enables multi-task learning for disease classification}. 
We describe these two salient phases, as well as random projection with attention noise mechanism in the following sections.

%\vspace{-0.3cm}

\subsection{Modality-Specific Task Learning Phase} \label{sec_MSTL}
\texttt{MSTL} consists of \( m \) modality-specific branches and incorporates two key blocks as follows:
%\vspace{-0.1cm}
\begin{itemize}[leftmargin=*]
    \item Efficient Residual Learning Attention (\texttt{ERLA}) block (Fig.~\ref{fig:fig3}) effectively learns refined multi-scaled diverse patterns — \( X^{'} = [x^{\prime}_{1}, \dots, x^{\prime}_{m}] \in \texttt{ERLA}(\cdot) \) for each modality-specific network branch.  
    \item Efficient Multimodal Cross Attention Module (\texttt{EMCAM}) block (Figs.~\ref{fig:fig3}-\ref{fig:fig4}), which takes \( x^{\prime}_{[1:m]} \) from each network branch as input to learn enhanced complementary shared representations \( X^S \). 
    %To achieve this, EMCAM comprises two attention modules:  a) the Efficient Dual-Branch Cross-Attention (EDCA) module, which learns enhanced modality-specific representations (Fig.~\ref{fig:fig2}), and  b) the Multimodal Frequency-Spatial Domain Fusion Attention (MFSA) module, which further effectively learns enhanced complementary shared representations \( X^S \) from input modalities \( X \) (Fig.~\ref{fig:fig3}). 
\end{itemize}

%\st{In the following, we will delve into a detailed discussion of the \texttt{MSTL} and \texttt{TMTL} phases (Fig.~\ref{fig:fig3}) focusing primarily on the key components of the \texttt{MSTL} phase, namely \texttt{ERLA} and \texttt{EMCAM}.}

\subsubsection{\textbf{Efficient Residual Learning Attention Block (ERLA)}} \label{erla}

\texttt{ERLA} block~(ref. Fig.~\ref{fig:fig3}(C)) is an extension of the Multi-Scale Convolution Block (\texttt{MSCB}) from \texttt{EMCAD} \cite{rahman2024emcad} as it is enhanced with our novel Efficient Multi-scale Information Learning Attention (\texttt{EMILA}) module~(ref. Fig.~\ref{fig:fig3}(D)). 
Our \texttt{ERLA} block integrates \texttt{EMILA}, ReLU (\( \texttt{R}(.) \)), batch normalization (\( \texttt{BN}(.) \)), and a \( 1 \times 1 \) group-point-wise convolution (\( \texttt{GPC}(.) \)). Given input \( x \), the \texttt{ERLA} recursively refines multi-scale representations via skip connections, enabling progressive updates to \( x \) for improved regularization:  
% ----------------------------------------
%\vspace{-0.2cm}
\begin{small}
\begin{equation}
\texttt{ERLA}(x) = \texttt{R} \left( X + \texttt{BN} \left( \texttt{GPC} \left( \texttt{EMILA} \left( \texttt{R} \left( \texttt{BN} \left( \texttt{EMILA} (x) \right) \right) \right) \right) \right) \right).
\end{equation}
%\vspace{-0.45cm}
\label{eq:eq1}
\end{small}
% ----------------------------------------
Let us discuss~\texttt{EMILA} in the following.
\texttt{EMILA} (ref. Fig.~\ref{fig:fig3} (D)) is inspired by \texttt{EMCAD}’s \texttt{MSDC} \cite{rahman2024emcad} but introduces a key enhancement: channel attention. It consists of three components: a) a multi-scale group with depth-wise convolutions (\texttt{MSGDC}) block~(ref. Fig.~\ref{fig:fig3}(G)), b) channel shuffle \cite{rahman2024emcad}, and c) channel attention to learn enhanced multi-scale representations. Let us discuss~\texttt{MSGDC} block first. 
Our \texttt{MSGDC} block consists of three parallel branches: the first branch employs a \( \texttt{GPC}(.) \) layer, while the other two  utilizes multi-scale depth-wise convolutions (\(\texttt{DWC}_{k}(\cdot)\)) at multiple scales $k \in \{3 \times 3, 5 \times 5\}$. 
These branches extract diverse patterns, which are then fused to capture multi-scale spatial details as \( x^{\prime} = \texttt{MSGDC}(\cdot) \):
%\vspace{-0.3cm}
\begin{small}
\begin{equation}
\texttt{MSGDC}(x) = \theta \left( \forall_{k \in \{3 \times 3, 5 \times 5\}}, \texttt{DWC}_k (x), \texttt{GPC} (x) \right).
\end{equation}
\label{eq:eq2}
\end{small}
Since depth-wise convolutions overlook relationships among channels, in~\texttt{EMILA}, channel shuffle is utilized to shuffle channels across groups to incorporate relationships among channels \cite{rahman2024emcad}. 
Therefore, final \texttt{EMILA} module can be formulated as:
%\vspace{-0.2cm}
\begin{small}
\begin{equation}
\texttt{EMILA}(x^{\prime}) = x + \texttt{GPC} \Big( \texttt{CA}\left( \texttt{MSGDC}(x)\right) \Big).
\end{equation}
\vspace{-0.5cm}
\label{eq:eq3}
\end{small}

Let us in the following explain the channel attention (\(\texttt{\texttt{CA}}(.)\)) block. The output of~\texttt{MSGDC} block is used as input to~\texttt{CA} block~(ref. Fig.~\ref{fig:fig3}(E)).
The block is motivated to \textit{refine modality-specific} representations, that first incorporates multi-perspective pooling operations -- global average pooling (\(\texttt{GAP}(.)\)), global max pooling (\(\texttt{GMP}(.)\)), and global min pooling (\(\texttt{GMN}(.)\)) to capture diverse channel dependencies.
%(ref. Fig.~\ref{fig:fig3} (E)). 
The resulting contexts undergo fusion (addition and
subtraction followed by addition) to enhance channel diversity and compressed through two fully connected layers (\(f(.)\)). A modulation process adaptively scales informative channels using learnable parameters ($\vartheta_x$), followed by a sigmoid activation (\(\sigma(.)\)) to generate channel attention maps (\(A_c\)). These maps dynamically recalibrate the input \(x\) through element-wise multiplication, yielding refined representations \((x_c = \texttt{CA}(x^{\prime}))\) for each modality \(m\) (ref. Eq. \ref{eq:eq3}).
Next, \( \texttt{GPC}(.) \) restores the original number of channels while encoding inter-channel dependencies. A skip connection recursively updates \( x \) with refined representations, further enhancing regularization. 
We formalize our~\texttt{CA} block as:

%\vspace{-0.35cm}
\begin{small}
\begin{equation}
\begin{aligned}
    \texttt{CA}(x^{\prime}) = x^{\prime} \times \sigma \Bigg( f\Big( 
    \big[ \sum_{p=1}^{3} G_p(x^{\prime}) \big]
    + \big[ \mu \big] 
    \Big) \times \vartheta_x \Bigg),
\end{aligned}
%\vspace{-0.2cm}
\label{eq:eq3}
\end{equation}
\end{small}
where $G_p \in [ \texttt{GMP}, \texttt{GAP}, \texttt{GMN}]$ and $\mu \in \texttt{GMP}(x^{\prime}) - \texttt{GAP}(x^{\prime}) - \texttt{GMN}(x^{\prime})$.

\begin{comment}
Our \texttt{EMILA} module is formulated as:
\vspace{-0.35cm}
\begin{footnotesize}
\label{eq:eq3}
\begin{equation}
\begin{aligned}
    \texttt{CA}(x^{\prime}) = x^{\prime} \times \sigma \Bigg( f_{[1:2]} \Big( 
    \big[ \sum_{p=1}^{3} G_p(x^{\prime}) \big]
    + \big[ \texttt{GMP}(x^{\prime}) - \texttt{GAP}(x^{\prime}) - \texttt{GMN}(x^{\prime}) \big] 
    \Big) \times \vartheta_x \Bigg)
\end{aligned}
\vspace{-0.2cm}
\end{equation}
\end{footnotesize}
where $G_p \in [ \texttt{GMP}, \texttt{GAP}, \texttt{GMN}]$.
\vspace{-0.2cm}
\begin{footnotesize}
\begin{equation}
\texttt{EMILA}(x^{\prime}) = x + \texttt{GPC} \Big( \texttt{CA}\left(MSGDC(x)\right) \Big)
\end{equation}
\vspace{-0.5cm}
\label{eq:eq4}
\end{footnotesize}
\end{comment}

%\begin{footnotesize}
%\begin{equation}
%\texttt{EMILA}(X) = X + \texttt{GPC}_2 \left( X_C \right)  
%\quad \text{\&} \quad  
%X_C = X' \times \sigma \left( f_2 \left( f_1 \left( \texttt{GMP}(X') + \texttt{GAP}(X') \right) \right) \right).
%\end{equation}
%\end{footnotesize}

%\noindent where \( X' \in \texttt{MSGDC}(X) \).
%\vspace{-0.1cm}
\subsubsection{\textbf{Efficient Multimodal Cross Attention Module (EMCAM)}}  
In this section, we discuss our proposed \texttt{EMCAM} block (ref. Fig.~\ref{fig:fig3}(F), and~Fig.~\ref{fig:fig4}), which aims to learn enhanced complementary shared representations \(X^{S}\) by refining the input \( X^{'} = [x^{'}_{1}, \dots, x^{'}_{m}] \) obtained from the \texttt{ERLA} block. Here, \(x^{'}_{i \in [1:m]} \in \mathbb{R}^{\texttt{H} \times \texttt{W} \times \texttt{C}} \) represents the $i$-th component of \( m \) input modalities, where \( \texttt{H} \), \( \texttt{W} \), and \( \texttt{C} \) denote height, width, and channels dimensions, respectively. By integrating global frequency-domain contexts with fine-grained spatial details, \texttt{EMCAM} enhances the model’s ability to capture a broader range of dependencies across frequency and spatial domains. This facilitates effective multimodal fusion, leading to a richer understanding of multimodal images. 
To achieve this, \texttt{EMCAM} integrates two parallel attention module: 
\begin{itemize}[leftmargin=*]
   \item Multimodal Frequency-domain Information Fusion Attention (\texttt{MFIFA}) (ref. Fig. \ref{fig:fig4} (A)) -- for capturing multi-frequency multimodal global contexts.
   \item Efficient Multimodal Spatial-domain Cross Attention (\texttt{EMSCA}) (Fig. \ref{fig:fig4} (B)) -- for refining multimodal spatial details with minimal cost.
\end{itemize}

\begin{algorithm}[H]
\renewcommand{\thealgorithm}{2}
\caption{: Adversarial Training with \texttt{RPAN}}
\begin{algorithmic}[1]
\setlength{\itemsep}{1pt}
\setlength{\topsep}{1pt}
\STATE \small{\textbf{Require:}  }
\STATE \small{\texttt{Robust-MAIL} network \( \mathcal{F}(\cdot) \) with learning parameter \( \beta \); }
\STATE \small{Number of random projection filters for \texttt{RPAN} layer} \( N_{r_\xi} \); 
\STATE Weight decay for \texttt{RPF} \( \varrho \);  
\STATE Number of \texttt{RPAN} layers \( \tau_{E/b} \) across the \( E \)-th total number of \texttt{ERLA} and \texttt{EMCAM} blocks and \( b \)-th branches;  
\STATE \texttt{CA} (\( \varsigma \)), \texttt{EMSCA} (\( \zeta \)), \texttt{MFIFA} (\( \Gamma \)), and modulated attention noise (\( \eta_l^{\prime} \));  
\STATE Perturbation size \( \epsilon \);  
\STATE Attack step size \( a \);  
\STATE Number of attack iterations \( k \);  
\STATE Training set \( \{X \in x_i, Y = \{y_1, \cdots, y_t\}\} \);
\STATE \textbf{Procedure:}
\WHILE{not converged}
    \STATE Sample a batch \( \{bX, bY\}_{i=1}^n \) from \( \{X, Y\} \);    
    \STATE \small{\textbf{Apply \texttt{RPF}s for Attack phase:}}
    \FOR{\( j = 1 \) to \( N_r \)}
        \STATE \small{Initialize random filters \( \mathcal{R}_j \sim \mathcal{N}(0, r^2) \), where \( r \in \sigma \);}
    \ENDFOR    
    \STATE \small{\textbf{Apply \texttt{RPAN} Layers for Attack phase:}}
    \FOR{each \( E/b \)}
        \STATE \small{Compute \texttt{RPAN} to learn enhanced noisy features:}
        \STATE $  \tau(X) = bX \times \left( 
        \sum_{\Phi \in \{\Gamma, \zeta\}} 
        \Phi \left( \varsigma \left( (\mathcal{R}_{1:N_{r_\xi}} * bX) \times \eta^{\prime}_{l} \right) * \mathcal{R}_{1:N_{r_\xi}} \right) \times \eta^{\prime}_{l} 
        \right)$
        \STATE $\min_{\beta} \max_{X^*} \left( \mathcal{L}_{\texttt{TMTL}}(\mathcal{F}(X^*, \beta), Y) 
        + \varrho \left( \| \mathcal{R}_{N_r+1:N} \| + \| \tau_{1:E/b} \| \right) \right) $
        \STATE $ \text{s.t.}  \|X^* - X\| \leq \epsilon$
        \STATE where \( * \) denotes convolution.
    \ENDFOR
    \STATE \textbf{Generate Adversarial Examples:}
    \STATE \small{Randomly initialize adversarial perturbation \( \delta \);}
    \FOR{\( i = 1 \) to \( k \)}
        \STATE $\delta \gets \delta + a \cdot \text{sign} \left(\nabla_{bX} \mathcal{L}_{\texttt{TMTL}}(\mathcal{F}_\beta(bX^{*}), bY)\right)$
        \STATE $bX^{*} \gets \text{Clip}^\epsilon_{bX}(bX + \delta)$
    \ENDFOR
    \STATE \textbf{Apply \texttt{RPAN} integrated by \texttt{RPF}s and modulated attention noise for Inference phase:}
    \FOR{each \( E/b \)}
        \FOR{\( j = 1 \) to \( N_r \)}
            \STATE Initialize random filters \( \mathcal{R}_j \sim \mathcal{N}(0, r^2) \);
        \STATE $ \tau(X) = bX \times \left( 
        \sum_{\Phi \in \{\Gamma, \zeta\}} 
        \Phi \left( \varsigma \left( (\mathcal{R}_{1:N_{r_\xi}} * bX) \times \eta^{\prime}_{l} \right) * \mathcal{R}_{1:N_{r_\xi}} \right) \times \eta^{\prime}_{l} 
        \right)$
        \STATE $\min_{\beta[I]} \max_{X^*} \Big( \mathcal{L}_{\texttt{TMTL}}(\mathcal{F}(X^*, \beta[A]), Y) + \varrho \big( \| \mathcal{R}_{\mathcal{N}_r+1:\mathcal{N}} \| + \| \tau_{1:E/b} \| \big) \Big) $
        \STATE $ \text{s.t.} \quad \|X^* - X\| \leq \epsilon$
        \ENDFOR
\ENDFOR
        \STATE \textbf{Adversarial Training Optimization:}
        \STATE $ \begin{aligned}
        \beta = \beta - \nabla_\beta 
        \Big( & \mathcal{L}_{\texttt{TMTL}}(\mathcal{F}(X^*, \beta), Y) + \varrho \Big( \|\mathcal{R}_{N_r+1}, \ldots, \mathcal{R}_N \| \\
        & + \|\tau_1, \ldots, \tau_{E/b} \| \Big) \Big)
        \end{aligned}$ 
\ENDWHILE
\end{algorithmic}
\label{algo:algo2}
\end{algorithm}

The outputs of these  blocks are fused and normalized via a sigmoid activation ($\sigma(\cdot)$) to generate multimodal attention maps \(A_m\). These maps adaptively recalibrate \(X'\) through element-wise multiplication and channel-wise trainable parameters \(\vartheta_m \in \mathbb{R}^C\), which dynamically weights modality-specific contributions. The enhanced shared representation \(X^S\) is computed as:  
\begin{small}
%\vspace{-.2cm}
\begin{equation}
\label{eq:eq13}
\small
X^{S} = X^{'} \times A_m \times \vartheta _{m}.
%\vspace{-.25cm}
\end{equation}
\end{small}
The parallel design of \texttt{MFIFA} and \texttt{EMSCA} blocks allows \texttt{EMCAM} to \textit{jointly optimize global frequency contexts and spatial dependencies while maintaining computational efficiency}.  
The following sections provide a detailed discussion of these attention modules.

\noindent \textbf{A. \textbf{Multimodal Frequency-domain Information Fusion Attention} (\texttt{MFIFA})} module is designed to learn multi-frequency multimodal global contexts, denoted as \(\omega^{mf}_{X^{e}}\), by analyzing frequency-based relationships to facilitate seamless feature alignment across modalities through the reformulation of the input \( X^{'}  \). 
\texttt{MFIFA} is illustrated in~Fig. \ref{fig:fig4} (A) and consists of three stages:

\textbf{Stage 1: Multimodal Feature Decomposition --} \label{DF}
Given input \(X^{'} \) from the \texttt{ERLA} block, it first applies the discrete cosine transform (\texttt{DCT}) to convert spatial representations into frequency-domain information. These contexts are then decomposed into diverse frequency segments -- low, high, and mean -- by applying three global pooling operations -- $\texttt{GMN}(.)$, $\texttt{GAP}(.)$, and $\texttt{GMP}(.)$ -- across m modalities:

\begin{itemize}[leftmargin=*]
   \item To capture diverse low-frequency components \(\omega_{X^{'}}^{lw}\), it applies two global pooling operations -- \(\texttt{GMN}(.)\) and \(\texttt{GAP}(.)\) -- to \(X^{'}\), obtaining low frequencies -- \(\omega_{X^{'}}^{lw1}\) and \(\omega_{X^{'}}^{lw2}\), which are then fused to form the diverse low-frequency component i.e. \(\omega_{X^{'}}^{lw}\):
\begin{small}
%\vspace{-.3cm}
\begin{equation}
\label{eq:eq8}
%\small
\omega_{X^{'}}^{lw1} = \texttt{GMN}(X^{'}) \quad \text{and} \quad
\omega_{X^{'}}^{lw2} = \texttt{GAP}(X^{'}), \quad \text{then} \quad \omega_{X^{'}}^{lw} = \omega_{X^{'}}^{lw1} + \omega_{X^{'}}^{lw2}.
%\vspace{-.1cm}
\end{equation}
\end{small}
%\vspace{-0.3cm}
\item To capture diverse high-frequency components i.e., $\omega_{X^{'}}^{h}$, it computes the residuals by subtracting the low-frequency terms from $X^{'}$, yielding high frequencies -- $\omega_{X^{'}}^{h1}$ and $\omega_{X^{'}}^{h2}$. Additionally. we apply $\texttt(.)$ on $X^{'}$ to learn another high frequency: $\omega_{X^{'}}^{h3}$. These high frequencies are fused to form high-frequency component i.e., $\omega_{X^{'}}^{h}$ as:
\begin{small}
%\vspace{-.2cm}
%\vspace{-.1cm}
\begin{equation}
\label{eq:eq9}
%\small
%\omega_m^{h} = \gamma(X^{''}) 
\omega_{X^{'}}^{h1} = X^{'} - \omega_{X^{'}}^{lw1}, \quad \omega_{X^{'}}^{h2} = X^{'} - \omega_{X^{'}}^{lw2},
\quad \omega_{X^{'}}^{h3} = \texttt{GMP}(X^{'}).
%\vspace{-.1cm}
\end{equation}
%\vspace{-.1cm}
\end{small}
\item  The mean-frequency component i.e., $\omega_{X^{'}}^{a}$ is obtained by computing the difference between the high- and low-frequency components: $\omega_{X^{'}}^{h}$ and $\omega_{X^{'}}^{lw}$:
\begin{small}
%\vspace{-.05cm}
\begin{equation}
\label{eq:eq10}
%\small
\omega_{X^{'}}^{h} = \omega_{X^{'}}^{h1} + \omega_{X^{'}}^{h2} + \omega_{X^{'}}^{h3},  
\quad
\omega_{X^{'}}^{a} = \omega_{X^{'}}^{h} - \omega_{X^{'}}^{lw}.
%\vspace{-.1cm}
\end{equation}
\end{small}
\end{itemize}

\textbf{ Stage 2: Modulation --} 
The resulting frequency components -- (\(\omega_{X^{\prime}}^{h}\), \(\omega_{X^{\prime}}^{lw}\), \(\omega_{X^{\prime}}^{a}\)) -- are modulated via learnable parameters \(\alpha_m\), \(\wp_m\), and \(\gamma_m\) through element-wise multiplication. This adaptively re-weights their contributions based on modality-specific importance.

\textbf{ Stage 3: Composition --} \label{composition}
The resulting components are fused across modalities to learn multi-frequency global contexts, denoted as \(\omega^{mf}_{X^{'}}\). Specifically, for each modality branch input (e.g., \(x^{'}_{i} \in X^{\prime}\)), the corresponding low-, high-, and mean-frequency components are hierarchically fused with those from other modality branches (e.g., \(x^{'}_{j}\) for \(j \in [m]\)). Finally, these components undergo further fusion to capture these global contexts, followed by a sigmoid activation ($\sigma$) to generate the frequency-domain attention map ($A_f = \texttt{MFIFA}(X^{\prime})$):
\begin{small}
%\vspace{-.2cm}
\begin{equation}
\label{eq:eq11}
%\small
A_f = \sigma \bigg( \sum_{i=1}^m \left( \alpha_i \times \omega_{x^{'}_i}^{lw} + \wp_i \times \omega_{x^{'}_i}^{h} + \gamma_i \times \omega_{x^{'}_i}^{a} \right) \bigg).
%\vspace{-.25cm}
\end{equation}
\end{small}

\noindent \textbf{B. \textbf{Efficient Multimodal Spatial-domain Cross Attention (\texttt{EMSCA})}} module is designed (ref. Fig. \ref{fig:fig4} (B)) to efficiently capture multimodal spatial details through multi-scale representations and cross-modal attention. In particular, \texttt{EMSCA} leverages  \texttt{MSGDC} from the \texttt{EMILA} module. As, 1×1 point-wise/depth-wise convolutions lacks spatial awareness and scale diversity,~\texttt{MSGDC} block in~\texttt{EMSCA} leverages multiple scales (e.g., 3×3, 5×5) within grouped channels to capture multi-scale spatial contexts at minimal computational cost. 
For each modality branch input $x^{'}_{i} \in X^{\prime}$ where $i \in [1, m])$, it computes:
\begin{small}
%\vspace{-0.2cm}
\begin{equation}
\label{eq:eq_12}
%\small
S(x^{\prime}_{i}) = \text{AP}(\texttt{MSGDC}(x^{\prime}_{i})) + \texttt{MP}(\texttt{MSGDC}(x^{\prime}_{i})),
\end{equation}
\end{small}
where average pooling (\texttt{AP}) and max pooling (\texttt{MP}) learn complementary spatial details. These learned patterns are fused recursively to progressively compress spatial dimensions, forming a \textit{hierarchical spatial refinement} as: 
\begin{small}
%\vspace{-0.2cm}
\begin{equation}
\label{eq:eq12}
\texttt{MSGDC}(x_{i}^{\prime}) \xrightarrow{\texttt{AP/MP}} \theta \xrightarrow{\text{fusion}} S(S(x_i^{'})).
\end{equation}
\end{small}
%\[ 
%\vspace{-0.3cm}
%\text{MSGDC}(X^e) \xrightarrow{\text{AP/MP}} F \xrightarrow{\text{fusion}} S(S(x_i^e)) 
%\]
To enable \textbf{cross-modal interaction} \label{CMI}, \texttt{EMSCA} employs symmetric skip connections between paired modalities (\( x_i^{'} \) and its counterpart \( x_{m-i+1}^{'} \)) to fuse base multimodal spatial representations \( (S(x_i^{'}) \) and \( S(x_{m-i+1}^{'})) \) along with hierarchical representations \( (S(S(x_i^{'}))) \), thereby learning cross-modal spatial contexts. This bidirectional interaction also enhances modality-specific regularization through recursive hierarchical compression. The learned spatial contexts are then adaptively modulated using learnable parameters \( \vartheta_i \), dynamically scaling each modality pair's contributions. Finally, feature fusion (addition) integrates the refined spatial details into the multimodal spatial context, followed by a sigmoid activation to generate spatial-domain attention map $A_s = \texttt{EMSCA}(\cdot)$: 
\begin{small}
%\vspace{-0.4cm}
\begin{equation}
\label{eq:eq12}
\texttt{EMSCA}(X^{\prime}) = \sum_{i=1}^{m} \vartheta_i \times \Big( \underbrace{S(x_i^{'}) + S(x_{m-i+1}^{'})}_{\text{cross-modal interaction}} + \underbrace{S(S(x_i^{'}))}_{\text{hierarchical representation}} \Big).
\end{equation}
\end{small}
%\vspace{-0.15cm}
Finally, to fuse the contributions of the \texttt{MFIFA} and \texttt{EMSCA} modules, we compute the final attention map \( A_m \) by adaptively scaling their outputs using learnable parameters \( \vartheta_f \) and \( \vartheta_s \). These parameters dynamically balance the relative importance of the two modules, enhancing overall performance. The attention map is computed as:
\begin{small}
%\vspace{-0.3cm}
\begin{equation}
\label{eq:eq13}
%\small
A_{m} = \sigma \left( \vartheta_{f} \times \texttt{MFIFA}({X^{\prime}}) + \vartheta_{s} \times \texttt{EMSCA}({X^{\prime}}) \right).
\end{equation}
%\vspace{-0.75cm}
\end{small}

% --------------------------------------------------------------------
\subsection{\textbf{Target-specific Multitask Learning}}
% --------------------------------------------------------------------

%follows \texttt{DRIFA-Net}'s MTL phase \cite{dhar2024multimodal}, 
The \texttt{TMTL} phase of~\texttt{MAIL} network uses shared representations \( X^S \) from  \texttt{MSTL} phase for multi-disease classification across \( m \) modalities.
It maps the input $X^S$ to task predictions \( Y \) through the loss function \( \mathcal{L}_{\texttt{TMTL}} \), where \( \lambda_t^m \) balances task-modality-specific losses \( \mathcal{L}_t^m \). The optimal model parameters \(\beta^*\) are learned by minimizing \( \mathcal{L}_{\texttt{TMTL}} \):
%\begin{footnotesize}
%\begin{equation} \label{eq:eq10} 
%\mathcal{L}_{\texttt{TMTL}} = \sum_{t} \lambda_t^m \cdot \mathcal{L}_t^m (F(X^S, Y)) \quad \text{and} \quad \beta^* = \arg \min_\beta  \left(\mathcal{L}_{\texttt{TMTL}})\right.  
%\end{equation}
%\end{footnotesize}

%\begin{equation} 
%\small
%\label{eq:eq10}
%\mathcal{L}_{\texttt{TMTL}} = \sum_{t=1}^{T} \sum_{m=1}^{M} \lambda_t^M \cdot \mathcal{L}_t^M \big( \mathcal{F}(X^S; \beta), Y \big) \quad \text{and} \quad \beta^* = \arg \min_\beta  \left(\mathcal{L}_{\texttt{TMTL}}), \right
%\end{equation}

\vspace{-0.3cm}
\begin{equation}
\small
\label{eq:eq10}
\mathcal{L}_{\texttt{TMTL}}
= \sum_{t=1}^{T} \sum_{m=1}^{M}
\lambda_{t}^{m}\,
\mathcal{L}_{t}^{m}\bigl(\mathcal{F}(X^{S};\beta),\,Y\bigr),
\quad
\beta^{*}
= \arg\min_{\beta}\,\mathcal{L}_{\texttt{TMTL}}.
\end{equation}

%\vspace{-0.05cm}
where $\beta$ signifies the \texttt{MAIL} parameters.
%\vspace{-0.1cm}
%\begin{equation} \label{eq:eq10}
 %   \partial_{\texttt{MTL}} = \sum_{t} \omega_t^m \times \partial_t^m (\theta(X^S, y_t)).
%\end{equation}

\begin{figure}[t]
    \centering
    \includegraphics[width=0.49\textwidth]{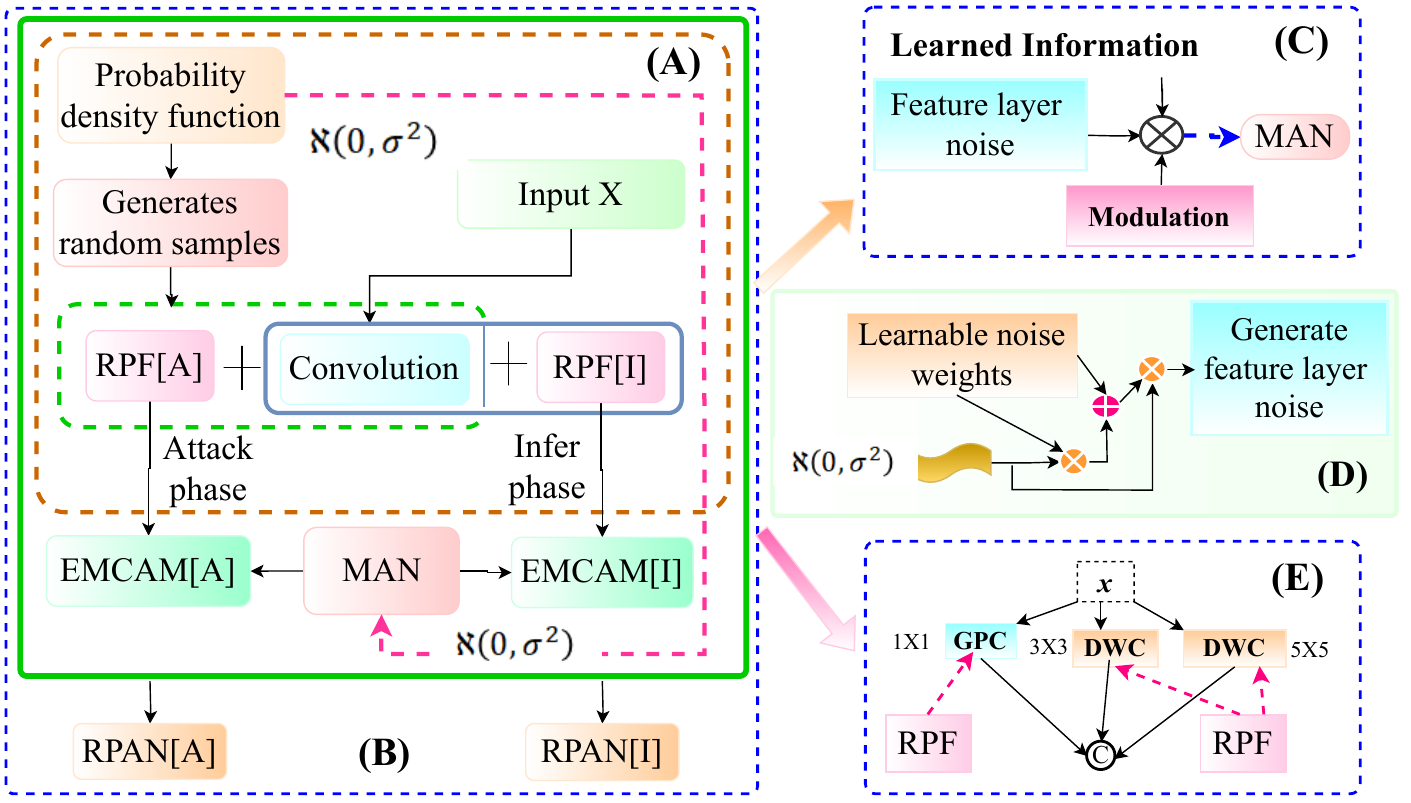}
        \vspace{-0.4cm}
    %\caption{\small \textbf{(A-B)} Overview of \texttt{Robust-MAIL} design, where \texttt{MAIL} incorporates \texttt{EMCAM} that integrates \texttt{RPF} and \texttt{MAN} are injected into \texttt{EMCAM} to form \texttt{RPAN} module. \texttt{RPAN} comprising key components: \texttt{RPF[A]} / \texttt{RPF[I]} (sampled Gaussian matrices), \texttt{EMCAM[A]} / \texttt{EMCAM[I]}, and \texttt{RPAN[A]} / \texttt{RPAN[I]} -- \texttt{[A]} and \texttt{[I]} shows attack and inference phases.
    %\textbf{(C)} shows how \texttt{MAN} is designed with leveraging learnable noises. \textbf{(D)} shows how \texttt{MAN} is integrated during modulation operations of \texttt{EMCAM} mechanism. \textbf{(E)} shows how \texttt{RPF} is integrated MSGDC component of \texttt{ERLA} and \texttt{EMCAM}  modules.}
    \caption{
    \small
    \textbf{(A-B)} Overview of \texttt{Robust-MAIL}, where \texttt{MAIL} integrates \texttt{EMCAM}, incorporating \texttt{RPF} and \texttt{MAN} to form the \texttt{RPAN} module. \texttt{RPAN} consists of key components: \texttt{RPF[A]} / \texttt{RPF[I]} (sampled Gaussian matrices), \texttt{EMCAM[A]} / \texttt{EMCAM[I]}, and \texttt{RPAN[A]} / \texttt{RPAN[I]}, where \textbf{[A]} and \textbf{[I]} denote attack and inference phases.  
    \textbf{(C–D)} Illustration of \texttt{MAN} utilizing learnable feature-layer noise, where noise is modulated through learnable weights combined with random noise to enhance robustness.  
    \textbf{(E)} Integration of \texttt{RPF} into the \texttt{MSGDC} from \texttt{ERLA} and \texttt{EMCAM}.
}
    \label{fig:fig5}
    %\vspace{-0.4cm}
\end{figure}

\begin{table*}[htbp]
\centering
\footnotesize
\caption{\small Performance comparison of \texttt{MAIL} with existing methods (\texttt{M1–M10}) on \texttt{D1–D10} datasets for classification. Bold and red indicate the best and second-best results, respectively. \texttt{MAIL} employs ResNet18 and ResNet50 as base networks, denoted as \texttt{MAIL-RN18} and \texttt{MAIL-RN50}.}
\vspace{-0.3cm}
%\textbf{(A)}
\label{tab:tab1}
\scalebox{0.808}{
\setlength{\tabcolsep}{2pt} % Adjust the column spacing (default is 6pt)
\begin{tabular}
{l|ccc|ccc|ccc|ccc|ccc|ccc|ccc|ccc|ccc|ccc|cc}
\hline \hline\noalign{\vskip 0.5pt}
\multirow{1}{*}{Datasets $\rightarrow$} & \multicolumn{3}{c|}{D1: Nickparvar} & \multicolumn{3}{c|}{D2: IQ-OTHNCCD} & \multicolumn{3}{c|}{D3: Tuberculosis} & \multicolumn{3}{c|}{D4: BCCD} & \multicolumn{3}{c|}{D5: HAM10000} & \multicolumn{3}{c|}{D6: SIPaKMeD} & \multicolumn{3}{c|}{D7: CRC}
& \multicolumn{3}{c|}{D8: CNMC-2019}
& \multicolumn{3}{c|}{D9: KVASIR}
& \multicolumn{3}{c|}{D10: CBIS-DDSM}
& \multicolumn{2}{c}{Cost Analysis}
\\ \hline \hline\noalign{\vskip 0.5pt}
 Models $\downarrow$ & \multicolumn{1}{c}{ACC} & F1 & AUC & \multicolumn{1}{c}{ACC} & F1 & AUC & \multicolumn{1}{c}{ACC} & F1 & AUC & \multicolumn{1}{c}{ACC} & F1 & AUC & \multicolumn{1}{c}{ACC} & F1 & AUC  & \multicolumn{1}{c}{ACC} & F1 & \multicolumn{1}{c|}{AUC}
 & \multicolumn{1}{c}{ACC} & F1 & \multicolumn{1}{c|}{AUC}
 & \multicolumn{1}{c}{ACC} & F1 & \multicolumn{1}{c|}{AUC}
 & \multicolumn{1}{c}{ACC} & F1 & \multicolumn{1}{c|}{AUC}
 & \multicolumn{1}{c}{ACC} & F1 & \multicolumn{1}{c|}{AUC} 
& \multicolumn{1}{c}{Params}
& \multicolumn{1}{c}{Flops}\\[0.5pt]
 \hline \hline\noalign{\vskip 0.5pt}
 \multirow{1}{*}{DDA-Net} & 96.8 & 96.6 & 97.1 
 & 98.1 & 97.4 & 98.1 & 
 98.6 & 97.7 & 98.8 & 
 98.2 & 98.2 & 98.6 
 & 92.2 & 91.8 & 92.6 
 & 92.5 & 92.8 & 91.9 
 & 93.4 & 92.7 & 93.6 
 & 92.6 & 91.8 & 93.1 
 & 90.6 & 90.2 & 91.1 
 & 92.5 & 91.8 & 91.5 
 & 12.1 & 1.12 \\
 \multirow{1}{*}{NAT} & 95.5 & 95.5 & 95.6 
& 97.5 & 97.3 & 97.2 
& 98.4 & 96.7 & 98.2 
& 98.1 & 97.7 & 98 
& 93.1 & 92.6 & 93.3 
& 91.1 & 91.1 & 91.5 
& 96.8 & 96.7 & 96.8 
& 94.1 & 93.7 & 93.4 
& 89.3 & 87.6 & 90.1 
& 92.0 & 91.7 & 91.9 
& 20 & 1.1 \\
\multirow{1}{*}{POTTER} & 95.1 & 94.7 & 94.7 
& 97.2 & 96.5 & 97.2 
& 97.8 & 97.2 & 98.2 
& 96.8 & 96.4 & 96.1 
& 91.3 & 91.1 & 91.8 
& 92.3 & 92.1 & 92.3 
& 95.8 & 95.2 & 96.2 
& 93.8 & 93.5 & 92.7 
& 91.5 & 90.7 & 90.9 
& 91.3 & 91.1 & 90.8
& 12 & 0.95 \\
\multirow{1}{*}{MFMSA} & 98.3 & 98.0 & 98.4 
& 99.5 & 99.3 & 99.5 & 
98.9 & 97.4 & 98.6 & 
98.3 & 98.5 & 98.8 & 
97.9 & 97.3 & 97.9 & 
94.7 & 94.4 & 95.3 & 
97.3 & 97.3 & 97.6 
& 96.2 & 96.0 & 96.3 
& 90.9 & 90.5 & 90.7 
& 94.6 & 94.5 & 94.6 
& 26.9 & 1.4 \\
\multirow{1}{*}{Gloria} & 98.1 & 97.5 & 97.9 
& 98.5 & 98.3 & 98.5 
& 98.2 & 97.5 & 98.9 
& 98.1 & 97.6 & 98.4 
& 93.7 & 93.7 & 94.5 
& 94.2 & 94.2 & 94.2 
& 96.2 & 95.5 & 95.9 
& 93.7 & 93.5 & 93.2 
& 89.6 & 89.1 & 90.2 
& 92.5 & 91.2 & 91.9 
& 30.8 & 1.54  \\
%\multirow{1}{*}{AsymFusion} 
%& 96.8 & 96.2 & 96.5 
%& 98.5 & 97.8 & 98.8 & 
%98.7 & 96.9 & 98.2 & 
%97.5 & 97.2 & 97.5 & 
%94.3 & 93.7 & 94.8 & 
%91.5 & 90.6 & 91.9 & 
%96.7 & 95.6 & 95.9 & 
%92.5 & 92.1 & 92.8 & 
%90.1 & 89.5 & 89.8 
%& 93.8 & 91.8 & 92.9
%& 118.2 & 16.1 \\
\multirow{1}{*}{MTTU-Net} & 
97.9 & 98.0 & 98.0 
& 99.5 & 99.2 & 99.5 
& 98.8 & 97.4 & 99 
& 97.8 & 97.6 & 98.2 
& 97.4 & 96.5 & 97.1 
& 91.9 & 92.3 & 92.5 
& 97.3 & 96.4 & 97.5 
& 94.3 & 93.6 & 94.2 
& 91.2 & 90.8 & 91.0 & 
94.1 & 93.3 & 94.5
& 38.1 & 6.8   \\
\multirow{1}{*}{HAMLET} & 96.2 & 95.9 & 96.2 
& 98 & 97.5 & 98.2 
& 97.4 & 97.1 & 97.1 
& 97.5 & 97.2 & 97.8 
& 93.5 & 93.4 & 93.2 
& 92.8 & 92.8 & 93.3 
& 96.7 & 96.3 & 97.2 
& 92.9 & 92.2 & 93.4 
& 89.3 & 87.4 & 88.6 
& 91.8 & 91.8 & 91.3 
& 57.3 & 3.52  \\
\multirow{1}{*}{MuMu} & 96.8 & 96.8 & 97.1 
& 98.2 & 97.9 & 98.7 
& 97.8 & 97.3 & 97.8 
& 97.9 & 97.5 & 97.8 
& 92.8 & 92.3 & 93.1 
& 92.3 & 91.7 & 92.8 
& 96.8 & 96.3 & 96.8 
& 93.3 & 92.7 & 93.8 
& 88.4 & 87.3 & 88.1 
& 92.1 & 91.7 & 92.5 
& 56.6 & 2.97  \\
\multirow{1}{*}{M$^3$Att} & 97.5 & 97.5 & 97.9 
& 98.8 & 98.8 & 98.8 
& 98.5 & 97.2 & 98.8 
& 98.3 & 98 & 98.5 
& 95.5 & 94.8 & 95.2 
& 92.2 & 91.4 & 92.3 
& 95.5 & 95.2 & 95.1 
& 93.9 & 93.1 & 94.2 
& 90.9 & 90.5 & 91.2 
& 93.2 & 92.7 & 93.6
& 183 & 12.14 \\
\multirow{1}{*}{DRIFA-Net} & 98.4 & 98.4 & 98.7 & 99.7 & 99.5 & 99.5 
& 99.1 & 97.4 & 99.1 
& 98.9 & 98.8 & 98.9 
& 98.2 & 97.9 & 98.5 
& 95.6 & 95.5 & 95.9 
& 97.5 & 97.2 & 97.6 
& 95.9 & 95.8 & 96.3 
& 91.9 & 91.1 & 91.7 
& 95.2 & 95.2 & 95.4
& 53.8 & 4.83 \\
\hline \hline\noalign{\vskip 0.5pt}
\multirow{1}{*}{MAIL-RN18} & \textcolor{red}{98.8} & \textcolor{red}{98.6} & \textcolor{red}{98.9} & \textcolor{red}{99.7} & \textcolor{red}{99.5} & \textcolor{red}{99.7} & 
 \textcolor{red}{99.3} & \textcolor{red}{97.8} & \textcolor{red}{99.3} & 
 \textcolor{red}{99.1} & \textcolor{red}{99.1} & \textcolor{red}{99} 
 & \textcolor{red}{99.4} & \textcolor{red}{99.4} & \textcolor{red}{99.7} 
 & \textcolor{red}{96.3} & \textcolor{red}{96.3} & \textcolor{red}{96.5} 
 & \textcolor{red}{98.4} & \textcolor{red}{97.9} & \textcolor{red}{98} 
 & \textcolor{red}{97.1} & \textcolor{red}{96.8} & \textcolor{red}{97.6} 
 & \textcolor{red}{92.3} & \textcolor{red}{91.6} & \textcolor{red}{92.3} 
 & \textcolor{red}{96.1} & \textcolor{red}{96.1} & \textcolor{red}{96.1}
 & 11.7 & 1.84 \\
\multirow{1}{*}{\cellcolor{orange!10}\textbf{MAIL-RN50}} & 
\cellcolor{orange!10}\textbf{99.2} & \cellcolor{orange!10}\textbf{99.1} & \cellcolor{orange!10}\textbf{99.2} & 
\cellcolor{orange!10}\textbf{99.7} & \cellcolor{orange!10}\textbf{99.5} & \cellcolor{orange!10}\textbf{99.5} & 
\cellcolor{orange!10}\textbf{99.5} & \cellcolor{orange!10}\textbf{98.1} & 
\cellcolor{orange!10}\textbf{99.4} & 
\cellcolor{orange!10}\textbf{99.3} & \cellcolor{orange!10}\textbf{99.3} & \cellcolor{orange!10}\textbf{99.6} & 
\cellcolor{orange!10}\textbf{99.8} & \cellcolor{orange!10}\textbf{99.6} & 
\cellcolor{orange!10}\textbf{99.8} & 
\cellcolor{orange!10} \textbf{98.1} & \cellcolor{orange!10}\textbf{98.1} & \cellcolor{orange!10}\textbf{98.9} & 
\cellcolor{orange!10}\textbf{98.7} & \cellcolor{orange!10}\textbf{98.3} & \cellcolor{orange!10}\textbf{98.7} & 
\cellcolor{orange!10}\textbf{97.3} & \cellcolor{orange!10}\textbf{97.1} & \cellcolor{orange!10}\textbf{97.8} & 
\cellcolor{orange!10}\textbf{92.7} & \cellcolor{orange!10}\textbf{92.2} & \cellcolor{orange!10}\textbf{93.5} & 
\cellcolor{orange!10}\textbf{96.5} & \cellcolor{orange!10}\textbf{96.5} & 
\cellcolor{orange!10}\textbf{96.9} & 
\cellcolor{orange!10}20.8 & 
\cellcolor{orange!10}2.18  \\
\hline \hline
\end{tabular}
}
%\vspace{-0.1cm}
\end{table*}

\begin{table*}[htbp]
\centering
\footnotesize
\caption{\small Performance comparison of \texttt{MAIL} with existing methods (\texttt{M1–M10}) on \texttt{D11–D16} (classification) and (\texttt{M11–M21}) on \texttt{D17–D20} (segmentation). Bold indicates the best result. \texttt{MAIL} employs MobileNetV2, ResNet18, and SegNet as base networks, denoted as \texttt{MAIL-M2}, \texttt{MAIL-RN18}, and \texttt{MAIL-Seg}.}
    \vspace{-0.3cm}
\label{tab:tab2}
\scalebox{0.75}{
\setlength{\tabcolsep}{2pt}
\begin{tabular}
{l|cc|cc|cc|cc|cc|cc||c|cc|cc|cc|cc|c|c}
\hline \hline\noalign{\vskip 0.5pt}
\multirow{1}{*}{Datasets $\rightarrow$} & \multicolumn{2}{c|}{D11: PathMNIST} & \multicolumn{2}{c|}{D12: PneumoniaMNIST} & \multicolumn{2}{c|}{D13: RetinaMNIST} & \multicolumn{2}{c|}{D14: BreastMNIST} & \multicolumn{2}{c|}{D15: TissueMNIST} & \multicolumn{2}{c||}{D16: OrganAMNIST} &  & \multicolumn{2}{c|}{D17: KVASIR} & \multicolumn{2}{c|}{D18: ISIC 2018} & \multicolumn{2}{c|}{D19: BraTs 2020} & \multicolumn{2}{c|}{D20: LiTs} & \multirow{2}{*}{Params} & \multirow{2}{*}{Flops}
\\[1pt] \cline{1-22}\noalign{\vskip 1pt}
 Models $\downarrow$ & \multicolumn{1}{c}{ACC} & AUC & \multicolumn{1}{c}{ACC} & AUC  & \multicolumn{1}{c}{ACC} & AUC & \multicolumn{1}{c}{ACC} & \multicolumn{1}{c|}{AUC} & \multicolumn{1}{c}{ACC} & \multicolumn{1}{c|}{AUC} & \multicolumn{1}{c}{ACC} & \multicolumn{1}{c||}{AUC} & \multirow{1}{*}{Models $\downarrow$} & \multicolumn{1}{c}{DICE} & \multicolumn{1}{c|}{mIOU} & \multicolumn{1}{c}{DICE} & \multicolumn{1}{c|}{mIOU} & \multicolumn{1}{c}{DICE} & \multicolumn{1}{c|}{mIOU} & \multicolumn{1}{c}{DICE} & \multicolumn{1}{c|}{mIOU} &  & \\[0.5pt]
 \hline \hline
 \multirow{1}{*}{DDA-Net} & 92.25 & 99.84 & 90.25 & 97.25 & 67.62 & 78.46 & 86.35 & 90.18 & 68.90 & 94.54 & 95.75 & 99.80 & UNet & 80.55 & 72.60 & 87.32 & 80.24 & 81.79 & 76.54 & 88.80 & 84.64 & 34.5 & 65.5
 \\
 \multirow{1}{*}{NAT} & 91.80 & 99.78 & 89.95 & 97.12 & 66.58 & 77.60 & 86.24 & 90.10 & 68.42 & 94.39 & 95.70 & 99.80 & UNet++ & 84.37 & 77.48 & 87.30 & 80.25 & 82.55 & 76.37 & 89.45 & 85.08 & 9.2 & 34.7
 \\
\multirow{1}{*}{POTTER} & 91.45 & 99.50 & 89.89 & 97.74 & 63.54 & 74.26 & 87.15 & 90.90 & 67.98 & 93.29 & 96.06 & 99.85 & AttnUNet & 83.89 & 77.15 & 87.80 & 80.54 & 82.78 & 76.48 & 86.72 & 84.98 & 34.9 & 66.6
\\
\multirow{1}{*}{MFMSA} & 92.28 & 99.70 & 91.49 & 98.15 & 69.10 & 78.84 & 88.90 & 93.37 & 72.45 & 95.29 & 96.12 & 99.90 & PolypPVT & 91.67 & 85.90 & 90.42 & 83.85 & 86.38 & 79.50 & 93.83 & 88.20 & 25.1 & 5.3
\\
\multirow{1}{*}{Gloria} & 91.94 & 99.78 & 91.25 & 97.88 & 66.89 & 77.55 & 87.52 & 91.07 & 69.88 & 94.72 & 95.68 & 99.80 & TransUNet & 86.45 & 81.37 & 87.28 & 81.20 & 84.90 & 78.24 & 92.12 & 87.85 & 105 & 38.5
\\
%\multirow{1}{*}{AsymFusion} & 91.52 & 99.25 & 89.93 & 97.78 & 67.42 & 77.25 & 87.92 & 91.16 & 69.84 & 93.87 & 95.50 & 99.80 & TransUNet & 86.45 & 81.37 & 87.28 & 81.20 & 84.90 & 78.24 & 92.12 & 87.85 & 105 & 38.5
%\\
\multirow{1}{*}{MTTU-Net} & 92.28 & 99.34 & 90.47 & 97.79 & 61.75 & 75.59 & 87.57 & 90.86 & 69.10 & 93.75 & 95.63 & 99.85 & SwinUNet & 88.50 & 82.74 & 88.77 & 81.93 & 85.56 & 78.80 & 93.90 & 88.32 & 27.2 & 6.2
\\
\multirow{1}{*}{HAMLET} & 91.85 & 99.29 & 89.42 & 97.65 & 63.38 & 75.81 & 86.68 & 90.48 & 69.58 & 93.98 & 95.28 & 99.80 & TransFuse & 88.95 & 83.50 & 89.14 & 82.31 & 83.80 & 77.25 & 91.34 & 86.25 & 143 & 82.7
\\
\multirow{1}{*}{MuMu} & 92.12 & 99.56 & 89.84 & 97.75 & 64.28 & 77.41 & 86.75 & 90.67 & 69.75 & 94.19 & 95.49 & 99.80 & MTTU-Net & 86.51 & 81.70 & 89.22 & 82.58 & 84.82 & 78.12 & 90.45 & 85.21 & 71.6 & 20.9 
\\
\multirow{1}{*}{M$^3$Att} & 91.72 & 99.15 & 91.25 & 97.82 & 63.15 & 76.90 & 87.84 & 91.55 & 70.54 & 94.52 & 95.84 & 99.85 & MADGNet & 90.77 & 85.38 & 90.20 & 83.76 & 90.18 & 81.32 & 94.35 & 89.14 & 31 & 14.2
\\
\multirow{1}{*}{DRIFA-Net} & 93.10 & 99.75 & 91.83 & 98.24 & 69.38 & 79.81 & 88.95 & 93.67 & 73.75 & 96.03 & 96.45 & 99.90 & DRIFA-Net & 90.90 & 86.14 & 90.55 & 83.87 & 89.92 & 81.25 & 94.50 & 89.82 & 67.3 & 19.94
\\

\multirow{1}{*}{MAIL-M2} & 93.25 & 99.84 & 92.05 & 98.60 & 69.45 & 80.54 & 89.65 & 93.80 & 74.18 & 96.27 & 96.79 & 99.90 & EMCAD & 92.84 & 86.45 & 90.91 & 84.12 & 90.53 & 82.55 & 94.57 & 89.82 & 26.8 & 5.6
\\

\multirow{1}{*}{\cellcolor{orange!10}\textbf{MAIL-RN18}} & 
\cellcolor{orange!10}\textbf{93.73} & \cellcolor{orange!10}\textbf{99.95} & \cellcolor{orange!10}\textbf{92.49} & \cellcolor{orange!10}\textbf{98.83} & \cellcolor{orange!10}\textbf{70.69} & \cellcolor{orange!10}\textbf{81.32} & \cellcolor{orange!10}\textbf{90.06} & \cellcolor{orange!10}\textbf{93.92} & \cellcolor{orange!10}\textbf{74.84} & \cellcolor{orange!10} \textbf{96.56} & \cellcolor{orange!10}\textbf{97.07} & \cellcolor{orange!10}\textbf{99.94} & \cellcolor{orange!10} \textbf{MAIL-Seg} & \cellcolor{orange!10}\textbf{93.05} & \cellcolor{orange!10}\textbf{86.50} & \cellcolor{orange!10}\textbf{91.27} & \cellcolor{orange!10}\textbf{84.46} & \cellcolor{orange!10}\textbf{91.13} & \cellcolor{orange!10}\textbf{82.78} & \cellcolor{orange!10}\textbf{95.16} & \cellcolor{orange!10}\textbf{90.27} & \cellcolor{orange!10}12.58 & \cellcolor{orange!10}7.89
\\
\hline \hline
\end{tabular}
}
%\vspace{-0.1cm}
\end{table*}

\begin{table*}[htbp]
\centering
\footnotesize
\caption{\small Comparison of \texttt{Robust-MAIL} with SOTA defenses (\texttt{M22–M27}) in accuracy on \texttt{D5}, \texttt{D3}, \texttt{D10}, \texttt{D11}, \texttt{D16}, and \texttt{D8} datasets. White-box (PGD, \texttt{BIM}, \texttt{MIM}) and black-box (\texttt{AutoAttack (AA)}, \texttt{Square}) attacks are used for evaluation. Bold indicates the best result.}
    \vspace{-0.3cm}
\setlength{\tabcolsep}{2.25pt}
\scalebox{0.78}{
\begin{tabular}
{l|ccccc|ccccc|ccccc|ccccc|ccccc|ccccc}
\hline \hline\noalign{\vskip 0.5pt}
\multirow{2}{*}{Datasets $\rightarrow$} & \multicolumn{5}{c|}{D5: HAM10000} & 
\multicolumn{5}{c|}{D3: Tuberculosis CXR} &
\multicolumn{5}{c|}{D10: CBIS-DDSM} &
\multicolumn{5}{c|}{D11: PathMNIST} & 
\multicolumn{5}{c|}{D16: OrganAMNIST} & 
\multicolumn{5}{c}{D8: CNMC-2019} \\
\cline{2-31}\noalign{\vskip 0.5pt} 
 & PGD & BIM & MIM & AA & Square & PGD & BIM & MIM & AA & Square & PGD & BIM & MIM & AA & Square & PGD & BIM & MIM & AA & Square & PGD & BIM & MIM & AA & Square & PGD & BIM & MIM & AA & Square \\ 
\hline \hline\noalign{\vskip 0.5pt}
PNI & 
51.8 & 55.2 & 53.5 & 60.9 & 63.1 & 
74.9 & 76.4 & 74.9 & 80.6 & 81.8 & 
47.6 & 49.4 & 47.3 & 59.2 & 56.2 & 
39.3 & 40.6 & 39.6 & 49.8 & 48.5 & 
39.8 & 38.1 & 37.4 & 49.4 & 46.8 & 
70.5 & 72.4 & 70.2 & 77.4 & 77.6 \\ 

DBN & 
48.3 & 50.9 & 48.3 & 57.6 & 59.4 & 
72.5 & 74.6 & 73.8 & 78.1 & 79.6 & 
40.3 & 45.9 & 40.3 & 58.4 & 56.2 & 
38.4 & 39.2 & 38.1 & 47.8 & 43.7 & 
38.7 & 39.5 & 38.0 & 45.6 & 45.3 & 
69.1 & 73.6 & 70.4 & 74.5 & 72.9 \\

Learn2Perturb &
53.0 & 54.6 & 51.8 & 61.4 & 59.4 & 
77.1 & 79.3 & 77.1 & 82.5 & 81.9 & 
45.8 & 43.2 & 45.8 & 61.1 & 62.3 & 
40.7 & 41.8 & 40.7 & 49.9 & 50.1 & 
39.4 & 44.7 & 39.4 & 48.2 & 47.8 & 
74.5 & 77.8 & 76.3 & 79.6 & 79.0 \\

RPF & 
59.5 & 62.1 & 60.3 & 68.9 & 69.4 & 
80.8 & 83.6 & 81.3 & 85.1 & 85.5 & 
51.0 & 49.3 & 48.0 & 67.2 & 63.7 & 
42.5 & 43.9 & 41.2 & 54.7 & 48.6 & 
41.8 & 40.6 & 38.8 & 52.9 & 51.2 & 
81.9 & 81.4 & 79.1 & 85.1 & 84.3 \\

CAP & 
60.8 & 63.2 & 56.4 & 69.2 & 72.4 & 
86.2 & 87.5 & 84.2 & 90.7 & 90.3 & 
52.3 & 52.2 & 50.1 & 68.1 & 64.5 & 
43.4 & 44.7 & 43.3 & 55.8 & 50.2 & 
40.7 & 42.1 & 41.2 & 51.9 & 50.6 & 
82.3 & 82.8 & 79.5 & 86.2 & 85.1 \\

CTRW & 
54.9 & 56.8 & 53.1 & 62.7 & 68.0 & 
82.4 & 84.3 & 82.4 & 87.1 & 88.1 & 
50.8 & 52.5 & 50.2 & 66.9 & 64.8 & 
41.9 & 40.2 & 39.1 & 52.5 & 50.7 & 
42.3 & 41.8 & 41.5 & 53.1 & 54.6 & 
80.8 & 78.8 & 75.8 & 84.6 & 84.3 \\
\hline \hline\noalign{\vskip 0.5pt}
\cellcolor{orange!10}\textbf{Robust-MAIL} & 
\cellcolor{orange!10} \textbf{65.6} & \cellcolor{orange!10} \textbf{66.3} & \cellcolor{orange!10} \textbf{65.8} & \cellcolor{orange!10} \textbf{78.6} & \cellcolor{orange!10} \textbf{80.7} & 
\cellcolor{orange!10} \textbf{90.5} & \cellcolor{orange!10} \textbf{89.7} & \cellcolor{orange!10} \textbf{86.5} & \cellcolor{orange!10} \textbf{92.3} & \cellcolor{orange!10} \textbf{91.4} & 
\cellcolor{orange!10} \textbf{54.3} & \cellcolor{orange!10} \textbf{54.7} & \cellcolor{orange!10} \textbf{52.3} & \cellcolor{orange!10} \textbf{70.5} & \cellcolor{orange!10} \textbf{66.4} & 
\cellcolor{orange!10} \textbf{45.7} & \cellcolor{orange!10} \textbf{45.7} & \cellcolor{orange!10} \textbf{44.5} & \cellcolor{orange!10} \textbf{58.3} & \cellcolor{orange!10} \textbf{54.2} & 
\cellcolor{orange!10} \textbf{44.2} & \cellcolor{orange!10} \textbf{43.0} & \cellcolor{orange!10} \textbf{42.9} & \cellcolor{orange!10} \textbf{55.6} & \cellcolor{orange!10} \textbf{56.1} & 
\cellcolor{orange!10} \textbf{84.4} & \cellcolor{orange!10} \textbf{84.7} & \cellcolor{orange!10} \textbf{82.2} & \cellcolor{orange!10} \textbf{87.5} & \cellcolor{orange!10} \textbf{87.8} \\
\hline \hline
\end{tabular}
}
\label{tab:tab3}
%\vspace{-0.22cm}
\end{table*}

\begin{figure*}[ht!]
    \centering
    \includegraphics[width=\textwidth]{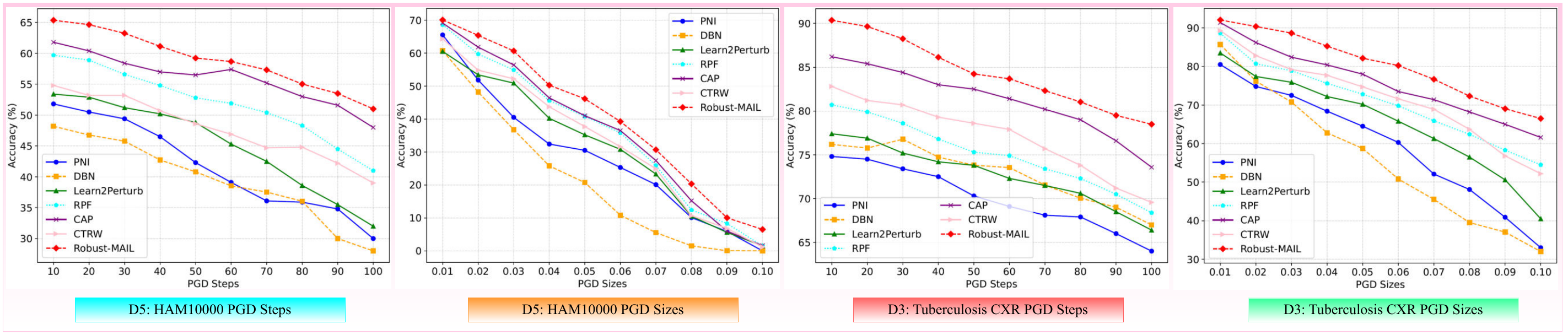}
        \vspace{-0.6cm}
    \caption{\small Evaluation of \texttt{Robust-MAIL}'s performance against stronger \texttt{PGD} attacks on the \texttt{D5} and \texttt{D3} datasets. %The x-axis corresponds to \texttt{PGD} steps and attack sizes.
    }
    \label{fig:fig6}
    %\vspace{-0.3cm}
\end{figure*}

\subsection{Random Projection with Attention Noise} \label{sec:RPF}

%Adversarial examples \(X^* \in X + \delta\), generated by adversarial attacks, disrupt efficient \texttt{MAIL} network's reliability by maximizing the loss \(\mathcal{L}_\texttt{{TMTL}}\), causing misclassifications via \(\mathcal{F}(X^*; \beta)\), where $\delta$ represents the perturbations:

%\begin{footnotesize}
%\vspace{-.2cm}
%\begin{equation} \label{eq:eq2}
%\small
%\max_{X^*} \quad \mathbb{E}_{(X, Y) \sim m} \left[\mathcal{L}_\texttt{{TMTL}}(\mathcal{F}(X^*; \beta), Y) \right] \; \text{s.t.} \; ||X^* - X|| \leq \epsilon.
%\vspace{-.1cm}
%\end{equation}
%\end{footnotesize}

Adversarial examples \( X^* = X + \delta \), generated by perturbing inputs \( X \) with bounded noise \( \delta \), compromise the reliability of the \texttt{MAIL} network by maximizing the loss \( \mathcal{L}_{\texttt{TMTL}} \). This causes misclassifications through the network’s predictions via \( \mathcal{F}(X^*; \beta) \), where \( \delta \) represents adversarial perturbations constrained by \( \|\delta\|_\infty \leq \epsilon \). The adversarial objective for modality \( m \) is formalized as:  
\begin{small}  
%\vspace{-.15cm}  
\begin{equation} \label{eq:eq_02}  
\max_{X^*} \; \mathbb{E}_{(X, Y) \sim \mathcal{D}_m} \left( \mathcal{L}_{\texttt{TMTL}} \big( \mathcal{F}(X^*; \beta), Y \big) \right) \quad \text{s.t.} \quad \|X^* - X\|_\infty \leq \epsilon,  
%\vspace{-.05cm}  
\end{equation}  
\end{small}  
where \( \mathcal{D}_m \) denotes the data distribution for modality \( m \). %, and \( \mathcal{L}_{\texttt{TMTL}} \) aggregates task-modality losses as defined in Eq.~\eqref{eq:eq10}. 
%To ensure reliability of the \texttt{MAIL} network against adversarial attacks, we extends the MAIL network by transforming into \texttt{Robust-MAIL}, which incorporates \textbf{R}andom \textbf{P}rojection combined with \textbf{A}ttention \textbf{N}oise (\texttt{RPAN}) mechanisms (see Fig. \ref{fig:fig2} (B)). \texttt{RPAN}, integrates \textbf{R}andom \textbf{P}rojection \textbf{F}ilter -- \texttt{RPF}, which is in form of sampled Gaussian matrices and learnable attention noise, integrated into the~\texttt{EMCAM} module. 
%as detailed in Section \ref{sec:RPF}. 
To ensure adversarial robustness in the \texttt{MAIL} network, we propose \texttt{Robust-MAIL}, an enhanced architecture incorporating \textit{Random Projection with Attention Noise} (\texttt{RPAN}) module (ref. Fig.~\ref{fig:fig5}). The \texttt{RPAN} module incorporates two key blocks within \texttt{EMCAM} block, described in the following:

\begin{itemize}[leftmargin=*] \item \textbf{Random Projection Filter (\texttt{RPF}):} Replaces conventional convolution filters in the depthwise (\texttt{DWC}) and point-wise (\texttt{GPC}) convolutions in \texttt{ERLA} and \texttt{EMCAM}, with randomly sampled Gaussian matrices (ref. Figs. \ref{fig:fig5} (A, E)). This introduces stochasticity into feature transformations, disrupting adversarial pattern propagation.

\item \textbf{Modulated Attention Noise (\texttt{MAN}):} Injects dynamically scaled learnable feature layer noise into \texttt{EMCAM}'s attention maps during modulation at training and inference (ref. Figs. \ref{fig:fig5} (A-D)). This adaptively corrupts adversarial gradients while smoothing learned representations.
%\vspace{-0.15cm}
\end{itemize}

\textbf{A. Random Projection Filter.} For a convolution layer with \( \mathcal{N} \) filters \( \mathcal{R}_{{1}}, \ldots, \mathcal{R}_\mathcal{N} \) %(\textcolor{green}{fix the range, starting from $\mathcal{N}_1$ to $\mathcal{N}$?})
, we partition them into two groups: (i) static randomly sampled \texttt{RPF}s $(\mathcal{R}_1, \ldots, \mathcal{R}_{\mathcal{N}_r}$ $\sim \mathcal{\eta}(0, \sigma^2))$ and (ii) trainable convolution filters \((\mathcal{R}_{\mathcal{N}_{r+1}},\) \(\ldots, \mathcal{R}_\mathcal{N})\) used in conventional, depthwise, and pointwise convolutions. These filters, with kernel size \( r \), are applied to the input \( X \).
Unlike prior works \cite{dong2023adversarial}, our approach uniquely integrates randomness across both depth-wise and group-point-wise convolutions in \texttt{ERLA} and \texttt{EMCAM} blocks (see Figure \ref{fig:fig5}(E)). 

Concurrently, modulated attention noise generates diverse noise patterns \(X^Z\), thereby further disrupting the generation of effective adversarial perturbations. 
Note that we use terms \texttt{[A]} and \texttt{[I]} to make explicit use of~\texttt{RPAN},~\texttt{RPF}, and~\texttt{EMCAM} in the attack and inference phases, respectively (ref. Figs. \ref{fig:fig5} (A-B)).   
By introducing stochasticity at both input and feature levels, \texttt{RPF} within \texttt{Robust-MAIL} significantly enhances adversarial defense. 

\textbf{B. Modulated Attention Noise.} \texttt{MAN} block is designed by injecting random learnable feature layer noise (\( \eta_l^{\prime} \)) (ref. Fig. \ref{fig:fig5} (D)) (alongside \texttt{RPF}s where applicable) into: a) \texttt{CA($\cdot$)} represented as (\( \varsigma \)), b) \texttt{MFIFA($\cdot$)} represented as (\( \Gamma \)) and c) \texttt{EMSCA($\cdot$)} represented as (\( \zeta \)), during their respective modulation processes. This enhances the learning of diverse noisy representations \( X^Z \), enhancing the adversarial robustness of the \texttt{MAIL} network. \emph{\textcolor{blue}{Additional details are provided in the \ref{sec:sec4} and \ref{sec:sec5}}.}

For adversarial robustness, \texttt{RPAN} is applied across \( E/b \) trainable convolution layers, where \( E \) is the total number of \texttt{ERLA} and \texttt{EMCAM} blocks and \( b \) indexes modality-specific branches (e.g., \( b = 2  \) ). \texttt{RPFs} apply across \( \mathcal{N}_{r_\xi} \)-th trainable layers in each \( (E/b) \), with \( \mathcal{N}_{r_\xi} \) denoting the total trainable convolutional layers per \texttt{EMCAM} module ($\xi$). The output enhanced noisy representations \( X^{Z} \) via the \texttt{RPAN} layer \( \tau(\cdot) \), for each $E$:
%\begin{footnotesize}
%\vspace{-.3cm}
%\begin{equation}
%\begin{aligned}
%X^{Z} &= X \times \Bigg( 
%\Gamma \Big( \varsigma \big( (\mathcal{R}_{1:N_{r_\xi}} * X) \times \eta^{'}_{l} \big) * \mathcal{R}_{1:N_{r_\xi}} \Big) \times \eta^{'}_{l} \\
%& \quad + \zeta \Big( \varsigma \big( (\mathcal{R}_{1:N_{r_\xi}} * X) \times \eta^{'}_{l} \big) * \mathcal{R}_{1:N_{r_\xi}} \Big) \times \eta^{'}_{l} 
%\Bigg)
%\end{aligned}
%\vspace{-.2cm}
%\end{equation}
%\end{footnotesize}
\begin{small}
%\vspace{-.4cm}
\begin{equation}
\begin{aligned}
\tau(X) &= X \times \bigg( 
\sum_{\Phi \in \{\Gamma, \zeta\}} 
\Phi \Big( \varsigma \big( (\mathcal{R}_{1:\mathcal{N}_{r_\xi}} * X) \times \eta^{\prime}_{l} \big) * \mathcal{R}_{1:\mathcal{N}_{r_\xi}} \Big) \times \eta^{\prime}_{l} 
\bigg).
\end{aligned}
%\vspace{-.1cm}
\end{equation}
\end{small}
The objective function for adversarial robustness in \texttt{Robust-MAIL} can be reformulated as:
%\begin{footnotesize}
%\vspace{-.1cm}
%\begin{equation}
%\footnotesize
%\small
%\begin{aligned}
%\min_{\beta} \max_{X^*} \left( \mathcal{L}_{\texttt{TMTL}}(\mathcal{F}(X^*, \beta), Y) + \varrho \left( \| \mathcal{R}_{N_r+1}, \ldots, \mathcal{R}_N \| + \right. \right. \\ 
%\left. \left.  \| \tau_1, \ldots, \tau_{E/b} \| \right) \right) \quad
%\text{s.t.} \quad \|X^* - X\| \leq \epsilon,
%\end{aligned}
%\vspace{-.2cm}
%\end{equation}
%\end{footnotesize}
\begin{small}
%\vspace{-.35cm}
\begin{equation} \label{eq_RPAN1_obj}
%\vspace{-.4cm}
\min_{\beta} \max_{X^*} \Big( \mathcal{L}_{\texttt{TMTL}}(\mathcal{F}(X^*, \beta), Y) + \varrho \big( \| \mathcal{R}_{\mathcal{N}_r+1:\mathcal{N}} \| + \| \tau_{1:E/b} \| \big) \Big)  
\ \text{} \text{s.t.} \ \|X^* - X\| \leq \epsilon.
%\vspace{-.05cm}
\end{equation}
\end{small}
where $\varrho$ denotes the hyperparameters of weight decay.

\subsubsection{\textbf{Adversarial Training with \textbf{\texttt{RPAN}}}} White-box attacks can efficiently craft perturbations \(\delta\) for a fixed network \(\mathcal{F}(\cdot)\) via gradient ascent. However, it is difficult for the generated adversarial example \(X^* \in X + \delta\) to successfully attack another network \(\mathcal{F}^{\prime}(\cdot)\) \cite{dong2023adversarial}.
%but the resulting adversarial example \(X^* \in X + \nu\) transfer poorly to black-box networks \(F'\) \cite{dong2023adversarial}. 
Following the \texttt{RPF} method, we generate adversarial examples using \texttt{RPAN}, denoted as \( \tau_{1:E/b}[A] \)  during the attack phase. These examples are then applied in adversarial training, where the \texttt{RPAN} layer \( \tau_{1:E/b}[I] \) is employed during inference (Fig. \ref{fig:fig5}). This setup reformulates the min-max optimization in Eq.~\ref{eq_RPAN1_obj} as:
\begin{small}
%\vspace{-.35cm}
\begin{equation}
\min_{\beta[I]} \max_{X^*} \Big( \mathcal{L}_{\texttt{TMTL}}(\mathcal{F}(X^*, \beta[A]), Y) + \varrho \big( \| \mathcal{R}_{\mathcal{N}_r+1:\mathcal{N}} \| + \| \tau_{1:E/b} \| \big) \Big).  
\end{equation}
\label{eq:eq18}
%\vspace{-.45cm}
\end{small}
\emph{Adversarial training with \texttt{RPAN} (\textcolor{blue}{ref. Algorithm \ref{algo:algo2}})  enhances \texttt{Robust}-\texttt{MAIL}’s adversarial robustness.} In white-box scenarios where attackers have access to \texttt{RPAN} parameters \( \tau_{1:E/b}[A] \), adversarial examples \( X^* \) are generated with respect to \( \mathcal{F}(X^*, \beta[A]) \). However, during evaluation, noise is regenerated by \( \tau_{1:E/b}[I] \), making \( X^* \) less vulnerable to \( \mathcal{F}(X^*, \beta[I]) \) and thus enhancing robustness against these attacks.

\begin{table}[t]
\centering
\caption{\small Performance comparison of \texttt{ERLA}, \texttt{MFIFA}, and \texttt{EMSCA} modules in \texttt{MAIL} on benchmark datasets \texttt{D5} and \texttt{D6}.}
    \vspace{-0.3cm}
\label{tab:tab4}
\setlength{\tabcolsep}{1pt}
\scalebox{0.66}{
\begin{tabular}{cccccc|cccccc|c}
\hline \hline\noalign{\vskip 2pt}
\textbf{Dataset} & \textbf{ERLA}  & \textbf{MFIFA}  & \textbf{EMSCA}  &  \textbf{Acc}   & \textbf{F1}     & \textbf{Dataset} & \textbf{ERLA}  & \textbf{MFIFA}  & \textbf{EMSCA}  &  \textbf{Acc}   & \textbf{F1}    & \textbf{Params} \\ \hline \hline\noalign{\vskip 2pt}
         \multirow{6}{*}{\textbf{D5}} &
         $\times$ & $\times$ & $\times$ &          94.9 & 94.7 &  
         \multirow{6}{*}{\textbf{D6}} & $\times$   & $\times$   & $\times$   & 90.1 & 88.9 & 23.4 \\ 
& 
         $\times$ & $\times$ & $\checkmark$ &          96.5 & 96.5  
&
& $\times$   & $\times$   & \checkmark   & 91.9 & 91.4 & 23.7 \\
& 
         $\times$ & $\checkmark$ & $\times$ &          96.9 & 96.8  
&
& $\times$   & $\checkmark$   & $\times$   & 92.3 & 92.3 & 23.9 \\
& 
         $\times$ & $\checkmark$ & $\checkmark$ &    \textcolor{red}{98.9} & \textcolor{red}{98.8}  
&
& $\times$   & $\checkmark$   & $\checkmark$   &  \textcolor{red}{95.8} &  \textcolor{red}{95.6} & 24.2 \\
& 
         $\checkmark$ & $\times$ & $\times$ &          95.9 & 95.6  
&
& $\checkmark$   & $\times$   & $\times$   & 91.5 & 91.2 & 9.83 \\
& 
         $\checkmark$ & $\times$ & $\checkmark$ &          97.7 & 97.7  
&
& $\checkmark$   & $\times$   & $\checkmark$   & 93.6 & 93.4 & 10.2 \\
& 
         $\checkmark$ & $\checkmark$ & $\times$ &          98.1 & 98.1  
&
& $\checkmark$   & $\checkmark$   & $\times$   & 94.5 & 94.5 & 10.4 \\
& 
         $\checkmark$ & $\checkmark$ & $\checkmark$ &        \cellcolor{orange!10} \textbf{99.4} & \cellcolor{orange!10} \textbf{99.4} 
&
& \checkmark   & \checkmark   & \checkmark   & \cellcolor{orange!10} \textbf{96.3} & \cellcolor{orange!10} \textbf{96.3} & 11.7 \\\noalign{\vskip 2pt}
\hline \hline
\end{tabular}}
%\vspace{-0.3cm}
\end{table}

\begin{table}[t]
\centering
\caption{\small 
Evaluation of EMILA components (MSGDC and CA), MFIFA components (CO and DF), and EMSCA components (MSGDC and CMI) (ref. section \ref{ablation}) on D5 and D6 datasets.
For \texttt{Robust-MAIL}, we assess the \texttt{RPAN} components: \texttt{RPF} and \texttt{MAN}, along with utilizing \texttt{RPF@E} on the \texttt{D5} and \texttt{D3} datasets.
"+" and "*" denote evaluations for \texttt{MAIL} and \texttt{Robust-MAIL}.
}
\vspace{-0.3cm}
\label{tab:tab5}
\setlength{\tabcolsep}{3pt}
\scalebox{0.595}{
\begin{tabular}{c|cc|cc|cc|c|c|c||c|ccc|c|c}
\hline \hline\noalign{\vskip 0.5pt}
& \multicolumn{2}{c}{\textbf{EMILA}} & \multicolumn{2}{|c}{\textbf{MFIFA}} & \multicolumn{2}{|c|}{\textbf{EMSCA}} & \textbf{D5} & \textbf{D6} &  & \multicolumn{4}{c|}{\textbf{Robust-MAIL}} & \textbf{D5} & \textbf{D3} \\
\hline\noalign{\vskip 2pt}
& MSGDC  & CA  & MO & DF  & MSGDC  & CMI  &  Acc   & Acc  & Params &  & RPF@E & RPF & MAN & Acc & Acc \\\noalign{\vskip 0.5pt} \hline   \hline\noalign{\vskip 0.5pt}
\multirow{9}{*}{+} &     $\times$ & $\times$ & $\times$ & $\times$ & $\times $ & $\times$
     & 94.9   & 90.1 & 23.4 & \multirow{9}{*}{*} &  $\times$ & $\times$ & $\times$ & 0 & 2.58 \\ 
&     $\checkmark$ & $\times$ & $\checkmark$ & $\times$  & $\checkmark$  & $\times$ & 96.49   & 92.15 & 9.85 & &  $\checkmark$ & $\times$ & $\times$ & 18.37 & 25.42  \\
&    $ \times$ & $\checkmark$ & $\times$ & $\checkmark$ & $\times$ & $\checkmark$ & 97.78   & 94.79 & 24.2 & &  $\times$ & $\times$ & $\checkmark$ & 48.27 & 73.39 \\
&     $\checkmark$ & $\times$ & $\times$ & $\checkmark$ & $\times$ & $\checkmark$ & 97.42   & 93.90 & 10.4 &  & $\times$ & $\checkmark$ & $\times$ & 46.95 & 70.44  \\
&     $\checkmark$ & $\checkmark$ & $\times$ & $\checkmark$ & $\times$ & $\checkmark$ & 98.48   & 95.52 & 10.8 & &  $\times$ & $\checkmark$ & $\checkmark$ & 61.18 & 82.56   \\
&     $\checkmark$ & $\checkmark$ & $\checkmark$ & $\times$  & $\checkmark$ & $\times$ & 97.51   & 94.47 & 10.2 & &  \textbf{---------} & \textbf{---------} & \textbf{---------} & \textbf{---------} & \textbf{---------}  \\
&     $\checkmark$ & $\checkmark$ & $\checkmark$ & $\checkmark$  & $\checkmark$ & $\times$ & \textcolor{red}{99.05}   & \textcolor{red}{96.10}  & 10.8 & & $\checkmark$ & $\times$ & $\checkmark$  & \textcolor{red}{63.29} & \textcolor{red}{86.50} \\
&     $\checkmark$ & $\checkmark$ & $\checkmark$ & $\times$ & $\checkmark$ & $\checkmark$ & 97.90   & 94.76 & 10.2 &  & $\checkmark$ & $\checkmark$ & x & 60.65 & 82.24  \\
&     $\checkmark$ & $\checkmark$ & $\checkmark$ & $\checkmark$ & $\checkmark$ & $\checkmark$ & \cellcolor{orange!10}\textbf{99.40}   & \cellcolor{orange!10}\textbf{96.31} & 11.7 &  & $\checkmark$ & $\checkmark$ & $\checkmark$ & \cellcolor{orange!10}\textbf{65.58} & \cellcolor{orange!10}\textbf{90.47}  \\\noalign{\vskip 0.5pt}
\hline \hline
\end{tabular}
}
%\vspace{-.3cm}
\end{table}

\begin{table}[t]
\centering
%\caption{Comparative evaluation of \texttt{ERLA},  \texttt{MFIFA} and \texttt{EMSCA} modules in \texttt{MAIL}, showing performance improvements on two benchmark datasets \texttt{D11} and \texttt{D19}.}
\caption{\small Performance comparison of cascaded attention (\texttt{CAT}) and parallel fusion attention (\texttt{PFA}) within our \texttt{MAIL} on \texttt{D5} and \texttt{D6} datasets.}
\vspace{-0.3cm}
\label{tab:tab6}
\setlength{\tabcolsep}{5pt}
\scalebox{0.66}{
\begin{tabular}{ccccc|ccccc|c}
\hline \hline\noalign{\vskip 2pt}
\textbf{Dataset} & \textbf{PFA}  & \textbf{CAT}  &  \textbf{Acc}   & \textbf{F1}     & \textbf{Dataset} &  \textbf{PFA}  & \textbf{CAT}  &  \textbf{Acc}   & \textbf{F1}    & \textbf{Params} \\ \hline \hline
         \multirow{2}{*}{\textbf{D5}} &
         $\times$ & $\checkmark$ &          99.1 & 99.0  
& \multirow{2}{*}{\textbf{D6}} & $\times$   & $\checkmark$   & 95.9 & 95.9 & 11.7 \\
& $\checkmark$ & $\times$ &        \cellcolor{orange!10} \textbf{99.4} & \cellcolor{orange!10} \textbf{99.4} 
&
&  $\checkmark$   & $\times$   & \cellcolor{orange!10} \textbf{96.3} & \cellcolor{orange!10} \textbf{96.3} & 11.7 \\\noalign{\vskip 2pt}
\hline \hline\noalign{\vskip 2pt}
\end{tabular}}
%\vspace{-0.5cm}
\end{table}

%-------------------------------------------------------------------------

\section{Experimental Analysis and Results}
\textbf{Datasets --} We utilized $20$ medical imaging datasets: \texttt{D1-D10} include \cite{nickparvar2021}, \cite{alyasriy2020iq}, \cite{rahman2020reliable}, \cite{ShengganBCCD}, \cite{tschandl2018ham10000}, \cite{plissiti2018sipakmed}, \cite{kather2016multi}, \cite{mourya2019all}, \cite{pogorelov2017kvasir}, and \cite{sawyer2016curated}; \texttt{D11-D16} are from \texttt{MedMNIST}; \texttt{D17-D20} comprise \cite{menze2014multimodal}, \cite{bilic2023liver}, \cite{jha2020kvasir}, and \cite{codella2019skin}. All images are resized to \(\small \texttt{128} \times \texttt{128} \times 3\) (for classification) and \(\small \texttt{224} \times \texttt{224} \times 3\) (for segmentation) with an 80\%/10\%/10\% train/val/test split (\emph{when applicable}) and standard augmentation applied. \emph{\textcolor{blue}{Dataset descriptions are provided in the \ref{sec:sec6}.}}

%%%%%%%%%%
\noindent\textbf{Data Preprocessing.} We augment each training image with three image-level transforms—rotation (20°), translation (5 px in both axes), and $3\times3$ Gaussian blur—producing three additional variants per sample. All augmented images inherit the original label. We then concatenate the augmented set with the original data (yielding a $4\times$ larger training set) and randomly shuffle the combined dataset to improve training diversity. These augmentations encourage robustness to common medical imaging variations in pose, spatial alignment, and acquisition noise.

%%%%%%%%%%%

\noindent \textbf{Models:} We evaluate \texttt{MAIL} against state-of-the-art (\texttt{SOTA}) single-modal learning and \texttt{MFL} networks across 16 classification and 4 segmentation datasets, with \texttt{Robust-MAIL} tested on 8 datasets for adversarial robustness.  
For classification, single-modal learning networks include \texttt{DDA}, \texttt{NAT}, \texttt{POTTER}, and \texttt{MFMSA} of \texttt{MADGNet} \cite{nam2024modality}~(denoted as~\texttt{M1-M4}) respectively, while \texttt{MFL} networks -- such as \texttt{Gloria} \cite{huang2021gloria}, 
%\texttt{MMTM} \cite{joze2020mmtm}, 
\texttt{MTTU-Net} \cite{cheng2022fully}, \texttt{HAMLET} \cite{islam2020hamlet}, \texttt{MuMu} \cite{islam2022mumu}, \texttt{M$^3$Att}, and \texttt{DRIFA-Net} \cite{dhar2024multimodal} -- are reconfigured to match our \texttt{MFL} network~(denoted as~\texttt{M5-M10}) respectively.  
For segmentation, we evaluate \texttt{MTTU}--\texttt{Net} and \texttt{DRIFA-Net} alongside \texttt{SOTA} models, including \texttt{UNet}\cite{ronneberger2015u}, \texttt{UNet++} \cite{zhou2018unet++}, \texttt{AttnUNet} \cite{oktay2018attention},  \texttt{PolypPVT} \cite{dong2021polyp}, \texttt{TransUNet} \cite{chen2021transunet}, \texttt{SwinUNet} \cite{cao2022swin}, \texttt{TransFuse} \cite{zhang2021transfuse}, \texttt{MADGNet} \cite{nam2024modality}, and \texttt{EMCAD} \cite{rahman2024emcad}~(denoted as~\texttt{M11}--\texttt{M21}) respectively.  
To assess adversarial robustness, we integrate methods like \texttt{PNI}, \texttt{DBN}, \texttt{Learn2Perturb}, \texttt{RPF}, \texttt{CAP}, and \texttt{CTRW} into \texttt{MAIL}~(denoted as~\texttt{M22-M27}) respectively.  
For classification (\texttt{D1-D16}), we use \texttt{ResNet18} \cite{he2016deep}, \texttt{ResNet50} \cite{he2016deep}, and \texttt{MobileNetV2} \cite{sandler2018mobilenetv2} as backbones, while \texttt{SegNet} \cite{badrinarayanan2017segnet} is used for segmentation (\texttt{D17-D20}).

\noindent \textbf{Notation --} Accuracy \texttt{(Acc)}, F1 score \texttt{(F1)}, and mean intersection over union \texttt{(mIOU)} are used for evaluation.

\begin{figure*}[ht!]
    \centering
    \includegraphics[width=0.75\textwidth,
    ]{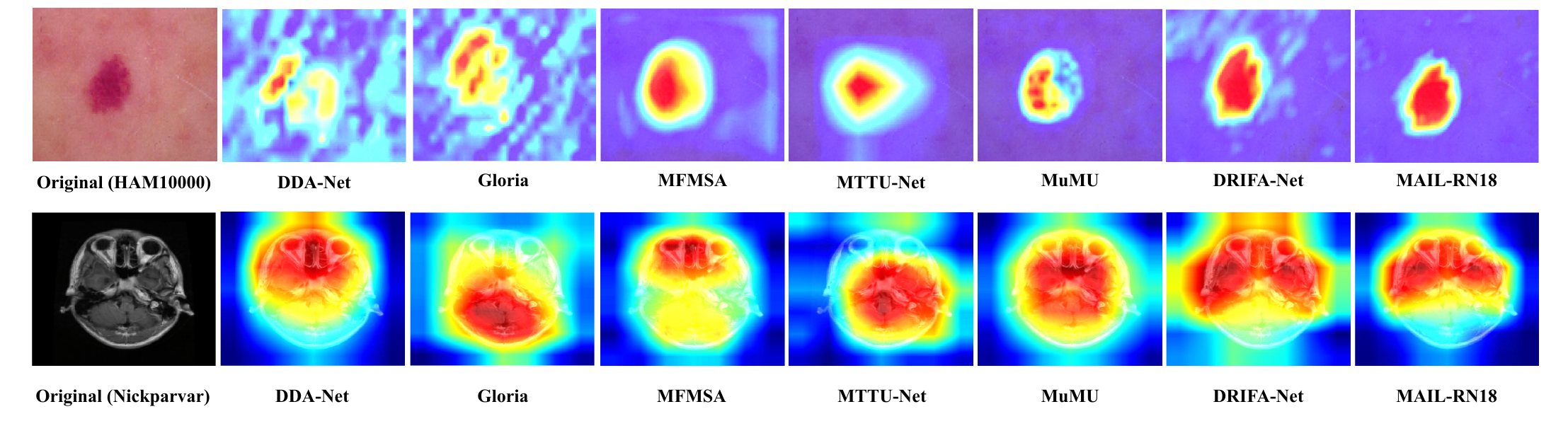}
    \vspace{-0.4cm}
    \caption{
    \small Visual representation of the important regions highlighted by our proposed \texttt{MAIL} networks i.e., \texttt{MAIL-RN18} and \texttt{MAIL-RN50} and ten \texttt{SOTA} methods using the
\texttt{GRAD-CAM} technique on two benchmark datasets \texttt{D5} and \texttt{D1}.
    }
    \label{fig:fig7}
    %\vspace{-0.4cm}
\end{figure*}

%%%%%%%%%%%%%%%%%%%%%%%%%%%%%%%%%%%%%%%%%%
%\begin{figure*}[ht!]
 %   \centering
  %  \includegraphics[width=1.0\textwidth,
   % height=0.2\textheight]{fig_7_acm_2.pdf}
   % \vspace{-0.7cm}
   % \caption{
   % \small T-SNE visualization comparing the proposed \texttt{MAIL-RN50} network with six \texttt{SOTA} methods on benchmark datasets \texttt{D11} and \texttt{D16}.
   % }
   % \label{fig:fig8}
   % \vspace{-0.4cm}
%\end{figure*}
%%%%%%%%%%%%%%%%%%%%%%%%%%%%%%%%%%%%%%%%%%%%%%%%%%

\noindent \textbf{Training Details --} All models were trained for 200 epochs using cross-entropy loss and the SGD optimizer (initial learning rate: 0.001) on an \texttt{NVIDIA RTX 4060 Ti GPU}. A ReduceLROnPlateau scheduler was used with a minimum learning rate of \(10^{-6}\). 
We follow the protocol of \texttt{SOTA} adversarial training strategy \cite{rice2020overfitting} to set up our experiments on our diverse datasets.

\noindent 
\textbf{Adversarial Evaluation --} \texttt{Robust-MAIL} is tested under white-box attacks \texttt{PGD} \cite{madry2017towards}, \texttt{BIM} \cite{kurakin2018adversarial}, and \texttt{MIM} \cite{dong2018boosting}, using \(\epsilon = \frac{4}{255}\), step size \(\alpha = \frac{10}{255}\), and 10–100 iterations. For black-box robustness, we adopt \texttt{AutoAttack (AA)} \cite{croce2020reliable} and \texttt{Square} \cite{andriushchenko2020square} with \texttt{ResNet101} as the surrogate. All experiments are implemented in \texttt{TensorFlow}. 
\subsection{Performance Comparisons}
%\vspace{-0.1cm}
In summary, our \texttt{MAIL} network achieved exceptional performance across 20 diverse medical imaging datasets, with classification (\texttt{D1-D16}) and segmentation (\texttt{D17-D20}) results between 70.69\% and 100\%. As shown in Tables \ref{tab:tab1}-\ref{tab:tab2}, \texttt{MAIL} surpassed \texttt{SOTA} single-modal learning and \texttt{MFL} methods with 0.2\%--13.9\% improvements across all metrics. We conducted a qualitative analysis on the \texttt{D1} and \texttt{D5} datasets further validate its effectiveness (ref. Figs. \ref{fig:fig7}).

As discussed earlier, \texttt{MAIL} outperforms prior methods due to their limited applicability across diverse medical imaging modalities. 
Although~\texttt{DRIFA-Net} performs well, it is resource-intensive attention modules incur high computational costs, limiting its applicability in resource-constrained environments. 
Our \texttt{MAIL} method \textcolor{blue}{addresses challenge 2} by parallelizing the fusion of the \texttt{MFIFA} and \texttt{EMSCA} modules, ensuring balanced contributions from both, thereby effectively capturing complementary shared representations \textcolor{blue}{(addressing challenge 3)} \cite{dhar2024multimodal}. By leveraging the \texttt{ERLA} and \texttt{EMCAM}, \texttt{MAIL} achieves optimal performance while minimizing computational costs. This design leads to performance gains of 0.2\%–1.9\% while significantly reducing computational cost -- achieving 54.9\%–81.3\% fewer parameters and FLOPs compared to top competitors \textcolor{blue}{(tackling challenge 1)} \cite{rahman2024emcad}.

%%%%%%%%%%%%%%
%Our \texttt{MAIL} network mitigates this through parallelized fusion of the \texttt{MFIFA} and \texttt{EMSCA} modules. By jointly integrating spatial and frequency domains, \texttt{MAIL} preserves richer multimodal information while ensuring balanced contributions from each module \textit{\textcolor{purple}{to address challenge 2}}. As discussed earlier, this design effectively learns complementary shared representations across diverse modalities \textit{\textcolor{purple}{to tackle challenge 3}}, improving performance by 0.1\%–1.6\% while reducing parameters and FLOPs by 54.9\%–78.3\%, thereby minimizing computational cost \textit{\textcolor{purple}{to address challenge 1}} \cite{rahman2024emcad}. 

%%%%%%%%%%%%%%%%%%%%%%%%%%%%%%%%%%%%%%%%%%%%%%%%%%%%%%%%%%%%%%%%%%%%%%%%%%%%%%%%%%%%%%%%%
%\vspace{-0.35cm}
\subsection{Impact of \textbf{\texttt{Robust-MAIL} Network}}\label{sub11}

As shown in Fig.~\ref{fig:fig1}, the baselines along with \texttt{MAIL} framework exhibits significant vulnerability to adversarial perturbations, underscoring the need for robust architectures to ensure reliability in AI-driven healthcare applications.  
To address this, we propose \texttt{Robust-MAIL}, which consistently outperforms \texttt{SOTA} defenses across six diverse datasets, achieving up to 9.34\% performance gains over leading competitors (Table \ref{tab:tab3}), \textcolor{blue}{addressing the limitations outlined in challenge 4}.  
Unlike existing methods that rely on a single defense approach, such as attention noise injection or \texttt{RPF}, our method uniquely integrates multiple defenses, including \texttt{MAN} and \texttt{RPF}, to achieve comprehensive robustness. 

\noindent\textbf{Evaluation with Stronger Attacks:} 
%Beyond the standard evaluation of adversarial robustness shown in Table \ref{tab:tab2} (B), 
We further evaluated \texttt{Robust}-\texttt{MAIL} on \texttt{D5} and \texttt{D3} datasets under stronger \texttt{PGD} attacks with increased iterations from 10 to 100 and larger perturbations $\epsilon \in [0.01, \cdots, 0.1]$ (Fig. \ref{fig:fig6}). Despite this situation, \texttt{Robust-MAIL} outperformed the best SOTA defenses, achieving up to 6.72\% higher performance under \texttt{PGD-100}. These results underscore \texttt{Robust-MAIL}’s superiority in protecting \texttt{MAIL} against stronger adversarial attacks.

%%%%%%%%%%%%%%%%

%\vspace{-0.3cm}
\subsection{Ablation Study}\label{ablation}
\textbf{Ablation of each component of~\texttt{MAIL}.} We evaluated the contribution of each component in \texttt{MAIL} on the \texttt{D5} and \texttt{D6} datasets, focusing on the \texttt{ERLA} block and the \texttt{MFIFA} and \texttt{EMSCA} modules of \texttt{EMCAM}. As shown in Table \ref{tab:tab4}, the \texttt{MAIL} network, incorporating all modules, outperformed variants missing any module, with gains ranging from $0.5\%$ to $7.4\%,$ highlighting the effectiveness of the integrated design.  

\noindent\textbf{Ablation of sub-components.}
We further analyzed individual contributions of \texttt{MSGDC} and \texttt{CA} in \texttt{EMILA}; the composition operation (\texttt{CO}) (ref. section \ref{composition}) and diverse frequency (\texttt{DF}) (ref. section \ref{DF}) -- in the \texttt{MFIFA}, and the role of \texttt{MSGDC} and cross-modal interaction \texttt{CMI}) in the \texttt{EMSCA} module. As shown in Table~\ref{tab:tab5}, \texttt{MAIL}, integrating all components, achieved performance gains of up to $6.2\%$ over partial configurations.  

\noindent\textbf{Ablation of cascaded vs. parallel fusion attention.}
Table \ref{tab:tab6} further shows that \texttt{MAIL}, with parallel fusion attention, outperforms cascaded attention by up to $0.4\%$, highlighting the benefit of parallel information integration.  

\noindent\textbf{Discussion.}
One can infer that the limited performance of ablated variants (Tables~\ref{tab:tab4}–\ref{tab:tab5}) can be attributed to their inability to capture effective complementary shared representations while maintaining low computational costs -- crucial for \textit{\textcolor{blue}{addressing challenge 1}}. 
%In relation to \textit{\textcolor{blue}{challenge 2}}, 
Also, cascaded attention architectures suffer from progressive information loss during inter-module transitions, primarily due to the absence of parallel fusion mechanisms among attention modules. This limitation restricts holistic information preservation, impedes effective representation learning, and ultimately degrades performance (ref. Table \ref{tab:tab6}) \cite{ lv2024lightweight, shen2021parallel}.
%As discussed earlier, \texttt{MAIL} tackles \textcolor{purple}{challenges 1--2} through \texttt{ERLA}, \texttt{MFIFA}, and \texttt{EMSCA}, enabling effective complementary shared representation learning by employing efficient residual learning (via \texttt{ERLA}) and parallel attention fusion across spatial and frequency domains (via \texttt{MFIFA} and \texttt{EMSCA}).
On the other hand,~\texttt{MAIL}'s design preserves holistic information across modules to mitigate information loss, thus leading to superior performance in multi-disease classification with minimal computational cost (Table \ref{tab:tab6}).
%To \textcolor{purple}{address challenge 4} and 
To demonstrate adversarial robustness, we evaluated \texttt{RPF} in the \texttt{ERLA} block (denoted as \texttt{RPF@E}) and at \texttt{EMCAM}, alongside \texttt{MAN} in \texttt{Robust-MAIL} on \texttt{D5} and \texttt{D3}. As reported in Table~\ref{tab:tab5}, \texttt{Robust-MAIL} outperforms all variants lacking any defense components, achieving improvements of up to $65\%$. These results highlight the crucial role of each defense module in enhancing robustness.
%\vspace{-0.46cm}

%%%%%%%%%%%%%%%%%%%%

%\vspace{-0.3cm}
\section{Conclusion} \label{sec:concl}
%\vspace{-0.05cm}
We propose \texttt{MAIL}, which uses efficient multimodal cross attention module to unlock the full potential of shared representation learning across diverse modalities for classification and segmentation tasks in medical imaging. To ensure adversarial robustness, we introduce \texttt{Robust-MAIL}. Extensive evaluations demonstrate that our approach effectively utilizes the synergy across modalities and tasks, learning effective shared representations for multi-disease classification while ensuring adversarial robustness. Future work will explore advanced multi-omics analysis and stronger adversarial defenses.
\clearpage

\bibliographystyle{ACM-Reference-Format}
%%\small \bibliography{samplebase}
\small %%% -*-BibTeX-*-

\appendix

%%%%%%%%%%%%%%%%%%%%%%%%%%%%%

%%%%%%%%%

\begin{table*}[t]
\centering
\caption{Summary of datasets used in our study.}
\small
\scalebox{0.88}{
\setlength{\tabcolsep}{5pt}
\begin{tabular}{p{2.7cm} p{1.9cm} p{1.8cm} p{4.0cm} p{1.2cm} p{2.1cm} p{3.8cm}}
\hline
\textbf{Dataset} & \textbf{Modality} & \textbf{Task} & \textbf{Classes / Labels} & \textbf{\#Samples} & \textbf{Split} & \textbf{Remarks} \\
\hline
Nickparvar MRI~\cite{nickparvar2021} & Brain MRI (T1--weighted) & Classification & Glioma (1621), Meningioma (1645), No Tumor (2000), Pituitary (1757) & 7023 & 80\%/10\%/10\% & Aggregated from Figshare, SARTAJ, BrH35 datasets. \\
IQ-OTHNCCD~\cite{alyasriy2020iq} & Chest CT & Classification & Normal, Benign, Malignant  & 1098 & 80\%/10\%/10\% & Lung nodule diagnosis. \\
Tuberculosis CXR~\cite{rahman2020reliable} & Chest X-ray & Classification & Normal vs Tuberculosis & 4200 & 80\%/10\%/10\% & Binary TB screening. \\
BCCD~\cite{ShengganBCCD} & Microscopy (blood smear) & Classification & Eosinophil, Lymphocyte, Monocyte, Neutrophil & 9957 & 80\%/10\%/10\% & Leukocyte subtype recognition. \\
HAM10000~\cite{tschandl2018ham10000} & Dermoscopy & Classification & 7 skin lesion categories & 10015 & 80\%/10\%/10\% & $>$50\% histopathology-confirmed labels. \\
SIPaKMeD~\cite{plissiti2018sipakmed} & Cytology & Classification & 5 epithelial cell types & 4049 & 80\%/10\%/10\% & Pap smear cell classification of cervical cancer. \\
CRC (Colorectal Histology MNIST)~\cite{kather2016multi} & Histology (H\&E patches) & Classification & 8 tissue classes (balanced) & 5000 & 80\%/10\%/10\% & Tissue micro-patches for histology typing. \\
C-NMC 2019~\cite{mourya2019all} & Microscopy (single-cell) & Classification & Malignant vs Healthy lymphocytes & 15114 & 10661/1867/2586 & -- \\
Kvasir~\cite{pogorelov2017kvasir} & GI endoscopy & Classification & 8 GI categories (e.g., ulcer, polyp) & 4000 & 80\%/10\%/10\% & Multi-class endoscopic findings. \\
CBIS-DDSM~\cite{sawyer2016curated} & Mammography & Classification & Normal vs Cancer & 10239 & 80\%/10\%/10\% & Curated; biopsy-verified labels. \\
PathMNIST (MedMNIST)~\cite{yang2023medmnist} & Histology patches & Classification & 9 tissue classes & 107180 & 89996/10004/7180 & -- \\
PneumoniaMNIST (MedMNIST)~\cite{yang2023medmnist} & Chest X-ray patches & Classification & Normal vs Pneumonia & 5856 & 4708/524/624 & -- \\
RetinaMNIST (MedMNIST)~\cite{yang2023medmnist} & Fundus patches & Classification & 5-grade diabetic retinopathy & 1600 & 1080/120/400 & -- \\
BreastMNIST (MedMNIST)~\cite{yang2023medmnist} & Breast ultrasound & Classification & Benign vs Malignant & 780 & 546/78/156 & -- \\
TissueMNIST (MedMNIST)~\cite{yang2023medmnist} & Single-cell images & Classification & 8 cell types & 236386 & 165466/23640/47280 & -- \\
OrganAMNIST (MedMNIST)~\cite{yang2023medmnist} & Abdominal CT slices & Classification & 11 organ classes & 58830 & 34561/6491/17778 & -- \\
Kvasir-SEG~\cite{jha2020kvasir} & GI endoscopy & Segmentation & Polyp mask (binary) & 1000 & 80\%/10\%/10\% & Resolutions 332$\times$487 to 1920$\times$1072; resized to $224\times224\times3$. \\
ISIC 2018 (seg)~\cite{codella2019skin} & Dermoscopy & Segmentation & Lesion mask (binary) & 2594 & 2075/519 & Resolutions 720$\times$540 to 6708$\times$4439; resized to $224\times224\times3$. \\
BraTS 2020~\cite{menze2014multimodal} & Brain MRI (FLAIR/T1/T1ce/T2) & Segmentation & WT/TC/ET tumor regions & 369 & 201/35/133 & Multi-modal volumes with tumor subregion annotations; processed to 224$\times$224$\times$3. \\
LiTS~\cite{bilic2023liver} & Abdominal CT & Segmentation & Liver + tumor masks & 201 & 131/70 & Multi-center; manual masks; processed to 224$\times224\times3$. \\
\hline
\end{tabular}}
\label{tab:datasets}
\vspace{-0.5cm}
\end{table*}

%%%%%%%%%%%%%%%%%%%%%%%%%%%%%

\vspace{-0.5cm}
\renewcommand{\thesection}{Appendix~A}
\section{RPF: Random Projection Filter} \label{sec:sec4}
\addcontentsline{toc}{section}{Appendix A: Random Projection Filter}

To improve the adversarial robustness of the proposed \texttt{MAIL} network, we introduce \texttt{Robust-MAIL} by incorporating Random Projection Filters (\texttt{RPF}) into the \texttt{MSGDC} blocks of the \texttt{EMILA} and \texttt{EMSCA} modules, following \cite{dong2023adversarial}. Concretely, we replace a subset of the standard learnable convolution kernels with fixed random projection filters. This randomization reduces an attacker’s ability—especially in white-box settings—to craft perturbations that reliably exploit learned filter structures. Nevertheless, aggressively increasing the proportion of random projections can inject substantial noise and hinder optimization.

To balance this trade-off, \texttt{Robust-MAIL} adopts a mixed design: only a fraction of convolution kernels in the \texttt{MSGDC} blocks are replaced by \texttt{RPF}s. In contrast, \texttt{RPAN} replaces all convolution kernels in each convolutional layer with random projection filters to further strengthen adversarial robustness.

Formally, consider an input feature map \( x_i \in \mathbb{R}^{n \times n \times d} = X \), where \( n \) and \( d \) denote the spatial resolution and the channel dimension, respectively. Let \( R \in \mathbb{R}^{r \times r \times d} \) be a random projection filter of a kernel size \( r \). The output feature map \( z \) is computed as:

\begin{equation}
\small
%\vspace{-0.4cm}
    z(p, q) =  R * [x]^{r}_{p, q} = \sum_{i=0}^{r-1} \sum_{j=0}^{r-1} \sum_{k=0}^{d-1} R(i, j, k) \cdot x(p+i, q+j, k) \ \text{s.t.}  \ \forall_{i}
\end{equation}

where \([x_i]^{r}_{p, q}\) denotes the \( r \times r \) subregion of \( x_i \), spanning rows \( p \) to \( p + r - 1 \) and columns \( q \) to \( q + r - 1 \), used for the convolution operation.

%%%%%%%%%%%%%%%%%%%%%%%%%%

%%%%%%%%%%%%%%%%%%%%%%

%%%%%%%%%%%%%%%%%%%%%%%%%%%%%%%%%%

%%%%%%%%%%%%%%%%%%%%%%%%%%%%%%%%%

%%%%%%%%%%%%-------------------------------

%%%%%%%%%%%%-------------------------------
\renewcommand{\thesection}{Appendix~B}
\section{Extending Details of RPAN} \label{sec:sec5}
\addcontentsline{toc}{section}{Appendix B: Extending Details of RPAN}

To ensure the adversarial robustness of the \texttt{MAIL} network and enable reliable predictions across diverse medical imaging modalities, we propose the \texttt{Robust-MAIL} network. This framework integrates the Random Projection with Attention Noise (\texttt{RPAN}) approach, which redesigns the \texttt{EMCAM} and \texttt{ERLA} blocks by incorporating Random Projection Filters (\texttt{RPF}s) and Modulated Attention Noise (\texttt{MAN}), thereby establishing a robust defense mechanism. The \texttt{RPAN} consists of two core steps:

\noindent\textbf{Redesigning the \texttt{ERLA} Block.}  
We redesign the group point-wise convolution (\texttt{GPC}) and depthwise convolution (\texttt{DWC}) layers in the \texttt{MSGDC} module of the \texttt{ERLA} block by replacing their convolution filters with random projection filters. Inspired by \cite{dong2023adversarial}, we substitute these filters with \texttt{RPF}s denoted as \( \mathcal{R}_1, \ldots, \mathcal{R}_{N_{r_\xi}} \) across the various scales of the \texttt{MSGDC}. This modification reformulates \texttt{MSGDC} within the \texttt{ERLA} block of the \texttt{MAIL} network, thereby yielding \texttt{Robust-MAIL} and transforming Eq.~(2) into:

%\vspace{-4mm}
%\begin{equation}
%\footnotesize
%\begin{split}
%\hat{l}_{b|m} &= \theta\Big(\theta\big(\delta_{1}(\eta_1), \phi_{1}(\eta_1)\big), 
%\theta\big(\delta_{q}(\eta_q), \phi_{q}(\eta_q)\big)\Big), \\
%\text{s.t.} \quad
%\eta_1 &\in \sigma_{1}(\psi_{1}(x_l|X^{'})), \quad 
%\eta_q \in \sigma_{q}\big(\psi_{q}(\theta(\delta_{1}(\eta_1), \phi_{1}(\eta_1)))\big).
%\end{split} \label{eq:eq17}
%\vspace{-1mm}
%\end{equation}

%where \( | \) denotes the selection of either \( x_l \) as the input for the multi-branch \texttt{MGA} or \( X^{'} \) for the multi-modal \texttt{MGA}. The parameters \( \eta_1 \) and \( \eta_q \) represent the learned local information through the \( q \)-th \( 1 \times 1 \) convolution layers \( \psi \), followed by a sigmoid activation \( \sigma \).

%\vspace{-0.4cm}
\vspace{-0.3cm}
\begin{small}
\begin{equation}
\texttt{MSGDC}(x) = \theta \left( \forall_{k \in \{3, 5\}}, \texttt{DWC}_k (\mathcal{R}_{1:N_{r_\xi}} * (x_i)), \texttt{GPC} (\mathcal{R}_{1:N_{r_\xi}} * (x_i)) \right) 
\end{equation}
\vspace{-0.4cm}
%\label{eq:eq2}
\end{small}

%where $\forall_{i \in [1:m]}$.

We further reformulate the channel attention maps \( A_c \) into attention noise maps by injecting modulated attention noise into the Channel Attention (\texttt{CA}) mechanism (refer to Section \ref{sec:RPF}). This is achieved by redesigning Eq. \ref{eq:eq3} into Eq. \ref{eq:eq12}. The resulting attention noise is then used to learn enhanced modality-specific noisy representations \( x^{\prime} \), thereby improving adversarial robustness. 

These modulated attention noise maps are generated by leveraging learnable feature layer noise through an iterative fusion strategy. This strategy integrates channel-wise noise tensor weights with trainable random noise (ref. Eq. \ref{eq:eq19}). It enables the effective and adversarially robust learning of rich noisy contexts tailored to each modality-specific pattern.

\begin{small}
%\label{eq:eq3}
\vspace{-0.3cm}
\begin{equation}
\begin{aligned}
    \texttt{CA}(x^{\prime}) = x^{\prime} \times \sigma  \Bigg( f_{[1:2]} \Big( 
    \big[ \sum_{p=1}^{3} G_p(x^{\prime}) \big]
    + \big[ \mu \big] 
    \Big) \times \vartheta_x  \times \eta_l^{\prime} \Bigg)
\end{aligned}
\end{equation}
\end{small}

where $\mu$ represents $\texttt{GMP}(x^{\prime}) - \texttt{GAP}(x^{\prime}) - \texttt{GMN}(x^{\prime})$.

In particular, to design the learnable feature layer noise—\(\eta_{I}\)—which is used to generate modulated attention noise for injection into the \texttt{ERLA} block, we employ an iterative refinement process on channel-wise noise tensor weights \(\delta_l \in \mathbb{R}^c\), along with random noise \(\eta_l\). This strategy enhances noise representation by guiding \(\eta_I\) to learn richer and more informative noisy patterns. During this process, \(\delta_l\) is first modulated by the corresponding random noise \(\eta_l\), followed by an element-wise multiplication with \(\delta_l\). The resulting intermediate noise is then further multiplied by \(\eta_l\), ultimately forming the learnable feature layer noise \(\eta_{I}\), as detailed in Eq. \ref{eq:eq19}.

\vspace{-3mm}
\begin{equation}
\label{eq:eq19}
\eta_{I} = \eta_l \times (\delta_l + (\eta_l \times \delta_l))
\end{equation}

%%%%%%%%%%%%%%%%%%%%%%%%%%%%%%%%%%%%%%%%%%%%%%%%%%

%%%%%%%%%%%%%%%%%%%%%%%%%%%%%%%%%%%%%%%%%%%%%%%%%%%%

\noindent\textbf{Redesigning the \texttt{EMCAM} Block.}  
We redesign the group-point-wise convolution (\texttt{GPC}) and depthwise convolution (\texttt{DWC}) of the \texttt{MSGDC} within the \texttt{EMSCA} module of the \texttt{EMCAM} block by replacing them with \texttt{RPF}s. Inspired by \cite{dong2023adversarial}, we substitute the convolution filters with random projection filters (\texttt{RPF}s), denoted as \( \mathcal{R}_1, \ldots, \mathcal{R}_{N_{r_\xi}} \), for each scale of the \texttt{MSGDC}. These \texttt{RPF}s reformulate the \texttt{MSGDC} in the \texttt{EMSCA} module of the \texttt{EMCAM} block within the \texttt{MAIL} network, enabling its transition into the \texttt{Robust-MAIL} framework. This modification effectively transforms Eq. \ref{eq:eq_12} into following:

\begin{small}
%\vspace{-0.4cm}
\begin{equation}
%\label{eq:eq32}
%\small
S(x^{\prime}_{i}) = \text{AP}(\texttt{MSGDC}(\mathcal{R}_{1:N_{r_\xi}} * (x^{\prime}_{i}))) + \texttt{MP}(\texttt{MSGDC}(\mathcal{R}_{1:N_{r_\xi}} * (x^{\prime}_{i}))) 
%\vspace{-0.3cm}
\end{equation}
\end{small}

We further reformulate the frequency-domain and spatial-domain attention maps, \( A_f \) and \( A_S \), from the \texttt{MFIFA} and \texttt{EMSCA} modules, respectively, into frequency-domain attention noise maps \( A_f \) and spatial-domain attention noise maps \( A_S \) by injecting the aforementioned modulated attention noise into the frequency-domain attention (\texttt{MFIFA}) and spatial-domain attention (\texttt{EMSCA}) mechanisms (see Section 3.1.2.(A-B)). This is accomplished by redesigning Eqs. \ref{eq:eq11} and \ref{eq:eq12} into the following:

\begin{small}
\vspace{-.2cm}
\begin{equation}
%\label{eq:eq11}
%\small
A_f = \sigma \bigg( \sum_{i=1}^m \bigg( \bigg (\eta_l^{\prime} \times \Big ( \alpha_i \times \omega_{x^{'}_i}^{lw} \Big) \bigg) + \bigg( \eta_l^{\prime} \times \Big ( \wp_i \times \omega_{x^{'}_i}^{h} \Big ) \bigg) + \bigg( \eta_l^{\prime} \times \Big ( \gamma_i \times \omega_{x^{'}_i}^{a} \Big) \bigg) \bigg) \bigg)
%\vspace{-.25cm}
\end{equation}
\end{small}

\begin{small}
\vspace{-0.4cm}
\begin{equation}
%\label{eq:eq12}
A_S = \sum_{i=1}^{m} \bigg (\eta_l^{\prime} \times \bigg ( \vartheta_i \times \Big( \underbrace{S(x_i^{'}) + S(x_{m-i+1}^{'})}_{\text{cross-modal interaction}} + \underbrace{S(S(x_i^{'}))}_{\text{hierarchical representation}} \Big) \bigg) \bigg)
\end{equation}
\end{small}
%\vspace{-0.15cm}

The resulting attention noise is then used to learn enhanced complementary shared noisy representations \( X^S \), thereby improving the adversarial robustness of our \texttt{MAIL} network, denoted by \( \mathcal{F}(\cdot) \).  
This strategy facilitates effective adversarially robust learning of rich shared noisy contexts.

We reformulate the multimodal attention maps \( A_m \) into multimodal attention noise maps \( A_{m} \) by incorporating modulated attention noise, achieved by redesigning Eq. \ref{eq:eq13} into the following: 

\begin{small}
\vspace{-0.3cm}
\begin{equation}
%\label{eq:eq12}
%\small
A_{m} = \sigma \bigg (\eta_l^{\prime} \times \Big (\vartheta_{f} \times \texttt{MFIFA}({X^{\prime}}) + \vartheta_{s} \times \texttt{EMSCA}({X^{\prime}}) \Big) \bigg) 
\end{equation}
\vspace{-0.3cm}
\end{small}

This attention noise is then injected into the attention map to learn enhanced complementary shared noisy representations \( X^{S} \in [x_1^{S}, \cdots, x_m^{S}] \).

%%%%%%%%%%%%%%%%%%%

\noindent\textbf{Adversarial Training with \texttt{RPAN}.} \label{sec:sec5}
Given a \texttt{Robust-MAIL} network \( \mathcal{F}(\cdot) \) with parameters \( \beta \), which maps multimodal inputs \( X \in x_{i \in [1:m]} \) to the logits \( \mathcal{F}(X, \beta) \in \mathbb{R}^C \), where \( C \) denotes the number of output classes, the adversarial example \( X^{*} = X + \nu \) is defined as:
\begin{equation}
\small
    \max_{X^{*}} \mathcal{L}_{\texttt{TMTL}}(\mathcal{F}(X^*, \beta[A]), Y), \quad \text{s.t.} \quad \|X^{*} - X\| \leq \epsilon,
\end{equation}

where \( \epsilon \) represents the maximum allowable perturbation size, and \( Y = [y_1, \dots, y_t] \) is the set of ground-truth labels for \( t \) classification tasks.
In the adversarial training strategy, adversarial examples are generated and fed into the \texttt{Robust-MAIL} network to formulate a min-max optimization problem as follows:

%%%%%%%%%%%%%%%%%%%%%

\vspace{-0.3cm}
\begin{equation}
\small
    \min_{\beta[I]} \max_{X^{*}} \mathcal{L}_{\texttt{TMTL}}(\mathcal{F}(X^*, \beta[A]), Y), \quad \text{s.t.} \quad \|X^{*} - X\| \leq \epsilon.
\end{equation}

With the incorporation of \texttt{RPF} and our modulated attention noise within \texttt{RPAN}, along with the adversarial training strategy described in Algorithm \ref{algo:algo2}, the \texttt{Robust-MAIL} model demonstrates strong defense capabilities during inference, as illustrated in Fig. \ref{fig:fig1}.

\renewcommand{\thesection}{Appendix~C}
\section{Datasets Details} \label{sec:sec6}
\addcontentsline{toc}{section}{Appendix C: Datasets Details}

Our study leverages 20 diverse medical imaging datasets, we provide the details in Table \ref{tab:datasets}:

%%%%%%%%%%

%%%%%%%%%%%%%%%%%%%%%%%%%%%%%%%%%%%%%%%%%%%%%%%%%%

%%%%%%%%%%%%%%%%%%%%%%%%%%%%%%%%%%%%%%%%%%%%%%%%%%%%%%%

\end{document}